\documentclass[pmlr,twocolumn,10pt]{jmlr} 

\usepackage{booktabs}
\usepackage{multirow}
\usepackage{tabularray}
\usepackage[table]{xcolor}

\usepackage{tikz}
\usepackage{amsmath,amssymb}
\usetikzlibrary{calc, matrix, positioning, arrows.meta, shapes.geometric, fit, backgrounds, calc, decorations.pathreplacing}

\usepackage{siunitx}
\usepackage{enumitem}

\theorembodyfont{\upshape}
\theoremheaderfont{\scshape}
\theorempostheader{:}
\theoremsep{\newline}

\usepackage{tikz}
\usetikzlibrary{arrows.meta, positioning, shapes.geometric, fit, calc}

\jmlrvolume{333}
\jmlryear{2026}
\jmlrsubmitted{LEAVE UNSET}
\jmlrpublished{LEAVE UNSET}
\jmlrworkshop{Conference on Health, Inference, and Learning (CHIL) 2026} 

\title[TEMPO: Transformers for Temporal Disease Progression]{TEMPO: Transformers for Temporal Disease Progression from Cross-Sectional Data}

\author{%
\Name{Hongtao Hao} \Email{hongtaoh@cs.wisc.edu}\\
\addr University of Wisconsin---Madison, USA
\AND
\Name{Joseph L. Austerweil} \Email{joseph.austerweil@gmail.com}\\
\addr Chiba Institute of Technology, Japan \& University of Wisconsin---Madison, USA 
\AND 
\Name{for the Alzheimer’s Disease Neuroimaging Initiative\textsuperscript{*}}
}

\begin{document}

\maketitle

\begingroup
\renewcommand\thefootnote{*}
\footnotetext{
Data used in preparation of this article were obtained from the Alzheimer’s Disease Neuroimaging Initiative (ADNI) database (\url{adni.loni.usc.edu}). As such, the investigators within the ADNI contributed to the design and implementation of ADNI and/or provided data but did not participate in analysis or writing of this report. A complete listing of ADNI
investigators can be found in Appendix \ref{apd:adniinfo}.}
\endgroup

\begin{abstract}
Event-Based Models (EBMs) infer biomarker progression from cross-sectional data but typically only as ordinal sequences and rely on rigid model assumptions. We propose \textsc{Tempo}, a Transformer architecture that learns both ordinal and continuous event sequences through simulation-based supervised learning. \textsc{Tempo} uses two Transformer modules: one treats biomarkers as tokens to infer event sequencing; the other treats patients as tokens, representing each by their per-biomarker abnormality profile, to infer patients' disease stages. On synthetic benchmarks, \textsc{Tempo} reduces normalized Kendall's Tau distance by 52.89\% and staging MAE by 25.33\% compared to state-of-the-art SA-EBM, with larger reductions in high-dimensional settings (58.88\% and 61.10\%). Applied to ADNI, \textsc{Tempo} recovers a biologically plausible Alzheimer's progression: early medial temporal atrophy, followed by amyloid accumulation and cognitive decline, and late-stage tau pathology with terminal acceleration of global neurodegeneration---broadly consistent with established disease models. \textsc{Tempo} also eliminates the need to derive custom inference algorithms and enables rapid empirical comparison of generative hypotheses.

\end{abstract}

\paragraph*{Data and Code Availability}
ADNI data can be requested through \url{https://adni.loni.usc.edu/data-samples/adni-data}. Reproducible codes for this study are available at \url{https://github.com/jpcca/tempo}.

\paragraph*{Institutional Review Board (IRB)}
The IRB at University of Wisconsin-Madison has reviewed and approved the research (\#2025-1254).

\section{Introduction}
\label{sec:intro}
Characterizing the progression of neurodegenerative diseases, such as Alzheimer's Disease (AD), is a cornerstone of modern clinical research. Accurate models of disease progression can support early diagnosis, prediction of clinical progression, and the design of therapeutic  interventions~\citep{jack2010hypothetical}. While longitudinal cohorts like ADNI~\citep{mueller2005alzheimer} and NACC~\citep{nacc_uds} provide invaluable data, collecting dense long-term measurements in neurodegenerative disease is expensive, logistically challenging, and often inconsistent over time because technologies and assessments change~\citep{young2024data}. Consequently, there is a sustained research effort to develop computational models capable of inferring temporal trajectories from purely cross-sectional snapshots.


To bridge this gap, Event-Based Models~\citep[EBM,][]{fonteijn2012event} pioneered estimating the latent temporal sequence by which a disease progresses through biomarkers. According to EBM, an \textit{event} is defined as the switch of a biomarker from a benign distribution to a pathological one.  While recent advancements have improved the robustness of these models~\citep{young2018uncovering, venkatraghavan2019disease, firth2020sequences, tandon2023sebm, wijeratne2024unscrambling, csparsimonious, hao2025stageaware}, EBMs still face critical limitations. 

First, most EBMs are restricted to ordinal sequencing. They provide the order of events but fail to estimate the actual temporal spacing between them. While continuous models like TEBM \citep{wijeratne2023temporal} exist, they typically rely on longitudinal data to anchor their timelines. Second, EBMs typically impose rigid constraints on disease logic, e.g., the assumption of \textit{irreversibility}, the requirement that progression follows a strict, one-way trajectory where subjects cannot revert to an earlier stage. If the underlying biological reality involves more complex dynamics, EBM performance may degrade. Third, while recent high-dimensional models~\citep{tandon2023sebm, wijeratne2024unscrambling} have addressed computational scaling, they have not been rigorously validated for robustness across varying generative frameworks. 

Deep learning and Transformers offer a potential alternative~\citep{lecun2015deep, vaswani2017attention}, but they require ground-truth progression labels that do not exist in real-world data. We only possess noisy clinical proxies, such as diagnosis labels (CN, MCI, AD), which are insufficient for capturing fine-grained latent disease progression over biomarkers. Consequently, most deep learning applications in this domain have focused on supervised label conversion \citep{hoang2023vision, wang2025deep} rather than the discovery of the underlying biological sequence.

In this work, we propose \textsc{Tempo} (\textbf{T}ransformer \textbf{E}stimation of \textbf{M}arkers' \textbf{P}rogression \textbf{O}rder). Through Simulation-Based Supervised Learning, \textsc{Tempo} enables continuous progression estimation from cross-sectional data and solves the label scarcity problem. Instead of manually engineering an inference algorithm for a single framework, we train a Transformer on a wide range of synthetic datasets generated by diverse pathological hypotheses formalized as generative processes, including frameworks for which no closed-form inference algorithm exists. By training on millions of simulated patients with known ground-truth sequences and disease stages, \textsc{Tempo} learns progression directly from cross-sectional observations. 



Our contributions are summarized as follows: (1) we propose a Transformer-based model that significantly outperforms the current state-of-the-art~\citep{hao2025stageaware}, reducing normalized Tau distance by 52.89\% and staging MAE by 25.33\% compared to SA-EBM in low-dimensional settings (12 biomarkers), with larger reductions at higher dimensionality (58.88\% and 61.10\%, respectively, at 100 biomarkers); (2) \textsc{Tempo} achieves the largest relative improvement in normalized staging error with dimensionality among all evaluated methods ($-43\%$: from $5.7\%$ at $B=12$ to $3.2\%$ at $B=100$), suggesting the patient-level attention mechanism actively benefits from more biomarker tokens; (3) we demonstrate that \textsc{Tempo} can reliably estimate continuous temporal spacing between events using only cross-sectional data, a task that remains a significant challenge for traditional EBMs; (4) we introduce a methodology where Transformers serve as proxies for complex likelihood functions, both eliminating the need for manual derivation of inference algorithms and enabling researchers to empirically compare diverse generative frameworks for clinical alignment; and (5) once trained, \textsc{Tempo} processes each test dataset in seconds, compared to 5--55 minutes for benchmark algorithms at 100 biomarkers, making large-scale hypothesis comparison practical.

\section{Related Work}
\label{sec:related_work}
Disease progression modeling encompasses a diverse range of computational objectives. To position \textsc{Tempo}, we categorize existing literature into four primary research areas.

\paragraph{Label Conversion and Clinical Prediction} 
A significant body of deep learning research focuses on supervised label conversion: predicting whether a patient will progress from cognitively normal (CN) to mild cognitive impairment (MCI) or AD within a specific timeframe \citep{al2025novel, almalki2025early, hoang2023vision, wang2025deep}. While these models are clinically valuable for risk stratification, they neither aim to recover the latent biological sequence of biomarker pathology, nor do they provide insight into the temporal spacing between specific physiological changes.

\paragraph{Pseudotime and Latent Staging} 
Inspired by biophysical models, the pseudotime method assigns each patient a latent score (typically 0 to 1) representing their position on a disease continuum \citep{agrawal2024b}. This approach focuses on patient staging, refining the granularity of an individual's disease state beyond coarse clinical labels. However, it describes \textit{where} a patient is, but not the temporal ordering of biomarker changes.

\paragraph{Ordinal Event-Based Modeling} 
The EBM framework aims to uncover the global \textbf{ordinal sequence} of biomarker pathology from cross-sectional data \citep{fonteijn2012event}. Numerous variants have been proposed to handle subtypes \citep{young2018uncovering, tandon2024s, hao2025bayesian}, high-dimensional scaling \citep{tandon2023sebm, wijeratne2024unscrambling}, non-Gaussian biomarker distributions \citep{firth2020sequences}, and uncertainty in patient stage distributions \citep{hao2025stageaware}. Despite their utility, these models rely on assumptions that may be biologically unrealistic. For instance, EBMs infer only the ordering of biomarker events without estimating the temporal spacing between them. Knowing that amyloid abnormality precedes neurodegeneration is insufficient if the years-long gap between them \citep{jack2010hypothetical} defines the therapeutic window. Another assumption, conditional independence among biomarkers given a participant's disease stage, yields a cleaner, more tractable likelihood function but oversimplifies reality, as biomarkers such as MMSE and ADAS13 are often highly correlated \citep{schmidt2016learning}. Finally, although EBMs assume irreversibility, studies have documented reversion of cognitive decline, both with \citep{bredesen2014reversal, du2018physical} and without intervention \citep{malek2016reversion}.

\paragraph{Continuous Temporal Spacing} 
The most detailed form of progression modeling seeks to recover the continuous sequence, where the gaps between biomarker events have real-world temporal meaning. Currently, models capable of continuous estimation, such as TEBM \citep{wijeratne2023temporal}, require longitudinal data to anchor their estimates of time. \textsc{Tempo} distinguishes itself from this category by estimating continuous temporal spacing between events using only cross-sectional data. This is achieved through simulation-based supervised learning: models are trained on synthetic cross-sectional data with ground-truth continuous event timelines, enabling recovery of these timelines from real cross-sectional clinical data.

\section{Methods}
\label{sec:methods}
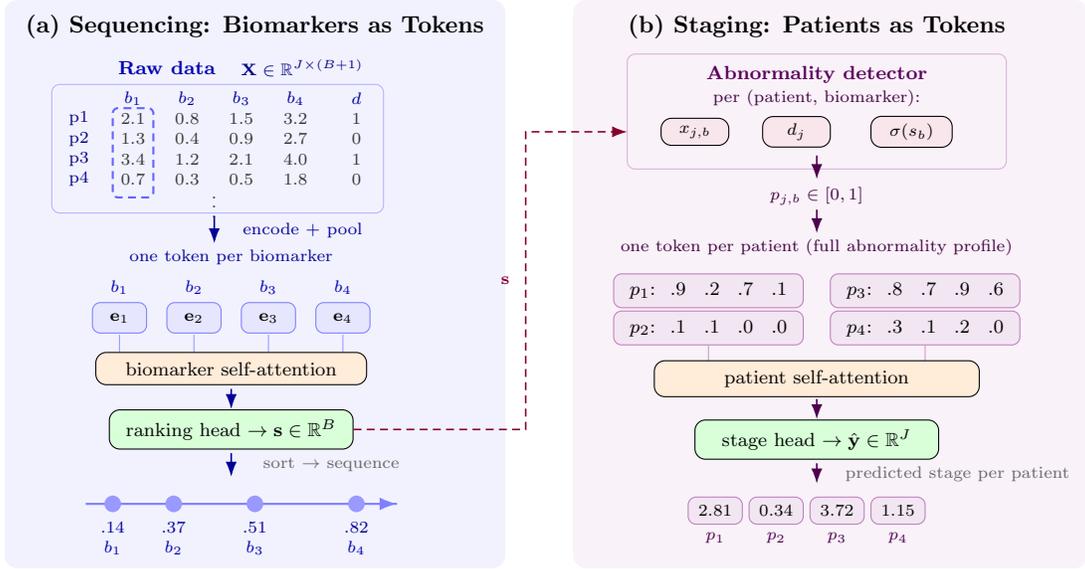
\begin{figure*}[t]
    \centering

\begin{tikzpicture}[scale=0.9, every node/.style={scale=0.9},
    font=\footnotesize,
    token/.style={draw, rounded corners=3pt, align=center, inner sep=2pt, minimum width=0.8cm, minimum height=0.6cm},
    bluetoken/.style={token, fill=blue!10, draw=blue!50},
    purptoken/.style={token, fill=violet!10, draw=violet!50},
    widetok/.style={draw, rounded corners=3pt, align=center, inner sep=3pt, minimum height=0.5cm, fill=violet!10, draw=violet!50, font=\footnotesize},
    attnbox/.style={draw, rounded corners, align=center, inner sep=4pt, minimum height=0.45cm, fill=orange!15, font=\footnotesize},
    outputbox/.style={draw, rounded corners, align=center, inner sep=4pt, minimum height=0.45cm, fill=green!15, font=\footnotesize},
    stagebox/.style={draw, rounded corners, align=center, inner sep=3pt, minimum height=0.4cm, fill=purple!10, font=\scriptsize},
    arrow/.style={-Latex, semithick},
    dashedarrow/.style={-Latex, semithick, densely dashed, purple!70!black},
    lbl/.style={font=\scriptsize},
    mathlbl/.style={font=\scriptsize, text=black!80},
    scorenode/.style={circle, fill=blue!40, inner sep=0pt, minimum size=7pt},
    ]

    \fill[blue!5, rounded corners=5pt] (-0.6,0.4) rectangle (6.8,8.8);
    \node[font=\normalsize\bfseries] at (3.1,8.4) {(a) Sequencing: Biomarkers as Tokens};

    \node[font=\footnotesize\bfseries, text=blue!70!black] at (1.8,7.8) {Raw data};
    \node[mathlbl, text=blue!50!black] at (3.8,7.8) {$\mathbf{X} \in \mathbb{R}^{J \times (B+1)}$};

    \node[mathlbl, text=blue!60!black] at (1.3,7.35) {$b_1$};
    \node[mathlbl, text=blue!60!black] at (2.1,7.35) {$b_2$};
    \node[mathlbl, text=blue!60!black] at (2.9,7.35) {$b_3$};
    \node[mathlbl, text=blue!60!black] at (3.7,7.35) {$b_4$};
    \node[mathlbl, text=blue!60!black] at (4.6,7.35) {$d$};

    \foreach \r/\lab/\va/\vb/\vc/\vd/\vdx in {
        7.05/p1/2.1/0.8/1.5/3.2/1,
        6.75/p2/1.3/0.4/0.9/2.7/0,
        6.45/p3/3.4/1.2/2.1/4.0/1,
        6.15/p4/0.7/0.3/0.5/1.8/0%
    }{
        \node[mathlbl, text=blue!50!black] at (0.5,\r) {\lab};
        \node[mathlbl] at (1.3,\r) {\va};
        \node[mathlbl] at (2.1,\r) {\vb};
        \node[mathlbl] at (2.9,\r) {\vc};
        \node[mathlbl] at (3.7,\r) {\vd};
        \node[mathlbl] at (4.6,\r) {\vdx};
    }
    \node[mathlbl] at (2.5,5.85) {$\vdots$};

    \draw[blue!30, rounded corners=3pt] (0.1,5.65) rectangle (5.0,7.55);
    \draw[blue!60, densely dashed, thick, rounded corners=2pt] (1.0,5.88) rectangle (1.6,7.2);

    \draw[arrow, blue!60!black] (2.5,5.6) -- (2.5,5.2);
    \node[mathlbl, text=blue!60!black, right] at (2.8,5.4) {encode + pool};

    \node[lbl, text=blue!60!black] at (2.75,5.0) {one token per biomarker};

    \node[mathlbl, text=blue!70!black] at (1.1,4.55) {$b_1$};
    \node[mathlbl, text=blue!70!black] at (2.2,4.55) {$b_2$};
    \node[mathlbl, text=blue!70!black] at (3.3,4.55) {$b_3$};
    \node[mathlbl, text=blue!70!black] at (4.4,4.55) {$b_4$};

    \node[bluetoken, minimum height=0.45cm] (e1) at (1.1,4.1) {\scriptsize$\mathbf{e}_1$};
    \node[bluetoken, minimum height=0.45cm] (e2) at (2.2,4.1) {\scriptsize$\mathbf{e}_2$};
    \node[bluetoken, minimum height=0.45cm] (e3) at (3.3,4.1) {\scriptsize$\mathbf{e}_3$};
    \node[bluetoken, minimum height=0.45cm] (e4) at (4.4,4.1) {\scriptsize$\mathbf{e}_4$};

    \foreach \x in {1.1,2.2,3.3,4.4}{
        \draw[blue!40] (\x,3.85) -- (\x,3.6);
    }

    \node[attnbox, minimum width=4.0cm] (battn) at (2.75,3.35) {biomarker self-attention};

    \draw[arrow, blue!60!black] (2.75,3.05) -- (2.75,2.75);
    \node[outputbox, minimum width=3.6cm] (rhead) at (2.75,2.45) {ranking head $\to \mathbf{s} \in \mathbb{R}^B$};

    \draw[arrow, blue!60!black] (2.75,2.1) -- (2.75,1.75);
    \node[lbl, text=black!60, right] at (3.1,1.93) {sort $\to$ sequence};

    \draw[blue!50, thick] (0.6,1.35) -- (5.2,1.35);
    \draw[-Latex, blue!50, thick] (5.0,1.35) -- (5.2,1.35);
    \node[scorenode] at (1.0,1.35) {};
    \node[scorenode] at (1.9,1.35) {};
    \node[scorenode] at (3.1,1.35) {};
    \node[scorenode] at (4.6,1.35) {};

    \node[mathlbl, text=blue!60!black] at (1.0,1.0) {.14};
    \node[mathlbl, text=blue!60!black] at (1.9,1.0) {.37};
    \node[mathlbl, text=blue!60!black] at (3.1,1.0) {.51};
    \node[mathlbl, text=blue!60!black] at (4.6,1.0) {.82};

    \node[mathlbl, text=blue!70!black] at (1.0,0.7) {$b_1$};
    \node[mathlbl, text=blue!70!black] at (1.9,0.7) {$b_2$};
    \node[mathlbl, text=blue!70!black] at (3.1,0.7) {$b_3$};
    \node[mathlbl, text=blue!70!black] at (4.6,0.7) {$b_4$};

    \fill[violet!5, rounded corners=5pt] (7.8,0.4) rectangle (15.4,8.8);
    \node[font=\normalsize\bfseries] at (11.4,8.4) {(b) Staging: Patients as Tokens};

    \draw[violet!30, rounded corners=3pt] (8.6,6.3) rectangle (14.2,8.0);
    \node[font=\footnotesize\bfseries, text=violet!70!black] at (11.4,7.7) {Abnormality detector};
    \node[lbl, text=violet!50!black] at (11.4,7.35) {per (patient, biomarker):};

    \node[stagebox, minimum width=1.0cm] at (9.6,6.85) {$x_{j,b}$};
    \node[stagebox, minimum width=1.0cm] at (11.1,6.85) {$d_j$};
    \node[stagebox, minimum width=1.2cm] at (12.8,6.85) {$\sigma(s_b)$};

    \draw[arrow, violet!60!black] (11.4,6.5) -- (11.4,6.15);
    \node[lbl, text=violet!50!black] at (11.4,5.9) {$p_{j,b} \in [0,1]$};

    \draw[dashedarrow] (rhead.east) -- (7.1,2.45) -- (7.1,6.85) -- (8.6,6.85);
    \node[lbl, text=purple!70!black] at (6.8,4.65) {$\mathbf{s}$};

    \draw[arrow, violet!60!black] (11.4,5.7) -- (11.4,5.35);
    \node[lbl, text=violet!50!black] at (11.4,5.15) {one token per patient (full abnormality profile)};

    \node[widetok, minimum width=2.8cm] (p1) at (9.8,4.5)
        {$p_1$:\; .9\;\; .2\;\; .7\;\; .1};
    \node[widetok, minimum width=2.8cm] (p2) at (9.8,3.95)
        {$p_2$:\; .1\;\; .1\;\; .0\;\; .0};
    \node[widetok, minimum width=2.8cm] (p3) at (13.0,4.5)
        {$p_3$:\; .8\;\; .7\;\; .9\;\; .6};
    \node[widetok, minimum width=2.8cm] (p4) at (13.0,3.95)
        {$p_4$:\; .3\;\; .1\;\; .2\;\; .0};

    \draw[violet!40] (9.8,3.7) -- (9.8,3.45);
    \draw[violet!40] (13.0,3.7) -- (13.0,3.45);

    \node[attnbox, minimum width=4.8cm] (pattn) at (11.4,3.2) {patient self-attention};

    \draw[arrow, violet!60!black] (11.4,2.9) -- (11.4,2.6);
    \node[outputbox, minimum width=3.6cm] (shead) at (11.4,2.3) {stage head $\to \hat{\mathbf{y}} \in \mathbb{R}^J$};

    \draw[arrow, violet!60!black] (11.4,1.95) -- (11.4,1.65);
    \node[lbl, text=black!60, right] at (11.7,1.8) {predicted stage per patient};

    \node[purptoken, minimum height=0.4cm] (y1) at (9.9,1.25) {\scriptsize 2.81};
    \node[purptoken, minimum height=0.4cm] (y2) at (10.8,1.25) {\scriptsize 0.34};
    \node[purptoken, minimum height=0.4cm] (y3) at (11.7,1.25) {\scriptsize 3.72};
    \node[purptoken, minimum height=0.4cm] (y4) at (12.6,1.25) {\scriptsize 1.15};

    \node[mathlbl, text=violet!60!black] at (9.9,0.85) {$p_1$};
    \node[mathlbl, text=violet!60!black] at (10.8,0.85) {$p_2$};
    \node[mathlbl, text=violet!60!black] at (11.7,0.85) {$p_3$};
    \node[mathlbl, text=violet!60!black] at (12.6,0.85) {$p_4$};

\end{tikzpicture}
    \caption{\textbf{TEMPO: Biomarkers and Patients as Tokens.}
(a)~\textbf{Sequencing:} Each biomarker's measurements across all $J$ patients are encoded and mean-pooled into a single token $\mathbf{e}_b$. Self-attention over biomarker tokens enables each to learn its relative cascade position. A ranking head outputs scores $\mathbf{s} \in \mathbb{R}^B$; sorting yields the event sequence.
(b)~\textbf{Staging:} Scores $\mathbf{s}$ are sigmoid-normalized and, with raw measurements $x_{j,b}$ and diagnosis labels $d_j$, fed into an \textit{Abnormality Detector} producing $p_{j,b} \in [0,1]$. Each patient's abnormality profile becomes a token. Cross-patient self-attention enriches representations with batch context, improving staging compared to MLP layers. A stage head predicts $\hat{y}_j$.
$B$: number of biomarkers; $J$: participants per batch; $B{+}1$ includes the diagnosis label ($d$). Refer to Fig.~\ref{fig:architecture} in Appendix~\ref{app:tempo_arch} for detailed \textsc{tempo} architecture.}
\label{fig:tokens}
\end{figure*}

We propose \textsc{Tempo} (Fig. \ref{fig:tokens}, and \ref{fig:architecture} in Appendix~\ref{app:tempo_arch}), a Transformer-based framework designed to simultaneously perform \textbf{event sequencing} and \textbf{patient staging} from cross-sectional data. To ensure numerical stability and allow for shared parameterization across disparate biomarker scales, all measurements are standardized (Z-score normalized) using statistics derived from the training population before being input into the model. Test data is normalized using the same statistics to prevent leakage.

\subsection{Event Sequencing Branch: Biomarkers-as-Tokens}
The sequencing branch treats each biomarker as a unique token to learn the global progression of pathology. Given a cohort of $J$ patients with $B$ biomarkers, let $\mathbf{X} \in \mathbb{R}^{J \times B}$ represent measurements and $\mathbf{d} \in \{0,1\}^J$ represent binary diagnosis labels (control vs. diseased).

\paragraph{Feature Extraction and Pooling.} To extract a global signature, i.e., a single cohort-level embedding, for each biomarker $b \in \{1, \dots, B\}$ we isolate its measurements across the entire cohort and concatenate them with the diagnosis labels, forming a local input $\mathbf{z}_b \in \mathbb{R}^{J \times 2}$. This input is processed by a \textit{Patient Encoder} $f_{\theta}$, a 2-layer Multi-Layer Perceptron (MLP) which maps each patient's biomarker measurement and diagnostic label to a latent space: $\mathbf{H}_b = f_{\theta}(\mathbf{z}_b) \in \mathbb{R}^{J \times d_{\text{model}}}$. We then apply mean pooling across the patient dimension to obtain a single $d_{\text{model}}$-dimensional embedding for that biomarker:

\begin{equation}
    \mathbf{e}_b = \frac{1}{J} \sum_{j=1}^{J} \mathbf{H}_{b, j} \in \mathbb{R}^{d_{\text{model}}}
\end{equation}

\noindent The resulting sequence of tokens $\mathbf{E} = [\mathbf{e}_1, \dots, \mathbf{e}_B]$ is augmented with learnable positional encodings before entering the Transformer.

\paragraph{Biomarker Transformer.} We employ a 4-layer Transformer Encoder to capture interactions between biomarkers. By computing multi-head self-attention on $\mathbf{E}$, the model learns the contextual relationships and pathological dependencies within the disease cascade. A final sequencing head projects the Transformer outputs to continuous event scores $\mathbf{s} \in \mathbb{R}^B$, which determine the relative order of biomarker abnormality. 


\subsection{Patient Staging Branch}

The staging branch consists of three components: an \textit{Abnormality Detector}, a \textit{Stage Encoder}, and a \textit{Stage Transformer}.

\paragraph{Abnormality Detector.} For each patient $j$ and biomarker $b$, a shared MLP takes three inputs: the standardized measurement $x_{j,b}$, the binary diagnosis label $d_j$, and the normalized ranking score $\sigma(s_b)$ (where $\sigma$ denotes the sigmoid function). It outputs a scalar abnormality probability:

$$
p_{j,b} = \text{AbnormalityDetector}(x_{j,b},\; d_j,\; \sigma(s_b)) \in [0, 1]
$$


Conditioning on the ranking score $s_b$ allows the detector to account for where biomarker $b$ falls in the estimated disease cascade when assessing its abnormality. The result is a matrix $\mathbf{P} \in [0,1]^{J \times B}$ of per-biomarker abnormality probabilities for each patient.

\paragraph{Stage Encoder and Patient Transformer.} Each patient's $B$-dimensional abnormality profile $\mathbf{p}_j = [p_{j,1}, \dots, p_{j,B}]$ is projected into a $2d_{\text{model}}$-dimensional latent space by the \textit{Stage Encoder} (a linear layer followed by layer normalization and ReLU). The resulting patient representations are treated as tokens and processed by a 2-layer Transformer Encoder, enabling patients to contextualize each other's abnormality profiles through cross-patient self-attention. Finally, a \textit{Stage Head} (a two-layer MLP with layer normalization) maps each patient's contextualized representation to a predicted stage $\hat{y}_j$.

\subsection{Loss Functions}
The model is trained using a multi-task loss $\mathcal{L} = \lambda_1 \mathcal{L}_{seq} + \lambda_2 \mathcal{L}_{stage}$ with $\lambda_1 = 1.0$ and $\lambda_2 = 1.0$.

\paragraph{Hybrid Sequencing Loss.} The sequencing branch is supervised by a dual-component loss: $\mathcal{L}_{seq} = 0.5 \mathcal{L}_{direct} + 0.5 \mathcal{L}_{pair}$. The \textbf{direct term} ($\mathcal{L}_{direct}$) anchors the predicted scores $\mathbf{s}$ to their absolute positions on the progression timeline. We compute the MSE between $\mathbf{s}$ and normalized ground-truth targets: discrete-order experiments use rank positions normalized to $[0, 1]$; continuous-order experiments use the raw event times, also normalized to $[0, 1]$. Although unbounded, $\mathbf{s}$ are implicitly regularized toward $[0, 1]$ by the MSE objective.

The \textbf{pairwise term} ($\mathcal{L}_{pair}$) explicitly supervises the relative ordering between biomarkers. At each training step, one valid pair $(a, b)$ is sampled, where valid means the two biomarkers occupy distinct positions in the ground-truth sequence ($r_a \neq r_b$). Let $s_a$ and $s_b$ denote the predicted scores for biomarkers $a$ and $b$. The form of $\mathcal{L}_{pair}$ depends on the nature of the ground truth.

For \textbf{discrete} event times, we treat ordering as a binary classification task:

\begin{align}
    \mathcal{L}_{pair}^{\text{discrete}} = -\big[ & y \log \sigma(s_b - s_a) \nonumber \\
    & + (1{-}y) \log (1 - \sigma(s_b - s_a)) \big]
\end{align}
where $y = \mathbf{1}[r_a < r_b]$. This is equivalent to maximum likelihood estimation under the Bradley--Terry model~\citep{bradley1952rank}, where $P(a \prec b) = \sigma(s_b - s_a)$.


For \textbf{continuous} event times, we treat ordering as a regression task:
\begin{equation}
    \mathcal{L}_{pair}^{\text{cont}} = \left( (s_b - s_a) - (t_b - t_a) \right)^2
\end{equation}
where $t_a, t_b \in [0,1]$ are the normalized ground-truth event times. Empirically, preserving actual temporal distances via MSE significantly outperformed converting them to binary labels for Binary Cross Entropy (BCE).



\paragraph{Normalized Staging Loss.} Patient staging is treated as a regression task using MSE. To ensure that the loss remains comparable across datasets with varying numbers of biomarkers, we normalize the mean squared error by the square of the biomarker count ($B^2$):
\begin{equation}
    \mathcal{L}_{stage} = \frac{1}{B^2} \left[ \frac{1}{J} \sum_{j=1}^J (\hat{y}_j - y_j^*)^2 \right]
    \label{eq:stage_loss}
\end{equation}
where $\hat{y}_j$ is the predicted stage and $y_j^*$ is the ground-truth \textbf{ordinal stage} (See Section~\ref{sec:benchmarking baseline}) for patient $j$.  This is equivalent to computing the loss on stages normalized to $[0, 1]$.

\section{Synthetic Experiments \& Results}
\label{sec:synthetic}

Unless otherwise noted, all metrics below are in-distribution: each model is evaluated on the 50 test datasets from its own experimental configuration. Cross-experiment results are reported separately in the Cross-Experiment Generalization paragraph.

\subsection{Low-Dimensional Experiments}
\label{sec:lowdim_exp}

We generated synthetic data following the framework in the SA-EBM study~\citep{hao2025stageaware} to evaluate \textsc{Tempo} under controlled conditions. The primary objective was to create synthetic cohorts that closely mirror the characteristics of the real-world ADNI dataset~\citep{mueller2005alzheimer}. To achieve this, we utilized the standard EBM algorithm \citep{fonteijn2012event, young2014data} to estimate the pre-event ($\phi$) and post-event ($\theta$) distribution parameters for 12 key biomarkers commonly reported in previous studies~\citep{csparsimonious, young2014data, archetti2019multi}. We adopted the data processing pipeline on ADNI from the SA-EBM work~\citep{hao2025stageaware}. See Section~\ref{sec:adni} and Appendix~\ref{apd:adni} for details on ADNI. 


\paragraph{Experiment Design.}
We conducted nine experiments (Fig.~\ref{fig:exp_design_matrix}; full details in Appendix~\ref{app:experiment_setup}), varying five generative dimensions: the type of biomarker event times ($\xi_b$: discrete integers vs.\ near-normal continuous), the patient stage type ($k_j$: ordinal vs.\ continuous), the stage prior (Dirichlet-Multinomial vs.\ Uniform for ordinal; Beta$(5,2)$ for continuous), the measurement model (EBM binary-switch vs.\ Sigmoid progression), and the biomarker distribution (Normal vs.\ Non-Normal for EBM measurement model). This design enables controlled comparisons: EBM vs.\ Sigmoid measurement (Exp.~5 vs.\ 6; Exp.~8 vs.\ 9), Normal vs.\ Non-Normal biomarkers (Exp.~1 vs.\ 2; Exp.~3 vs.\ 4; Exp.~6 vs.\ 7), DM vs.\ Uniform stage prior (Exp.~1 vs.\ 3; Exp.~2 vs.\ 4), and discrete vs.\ continuous event times (Exp.~5 vs.\ 8; Exp.~6 vs.\ 9).

\begin{figure}[htbp]
    \centering \includegraphics[width=0.45\textwidth]{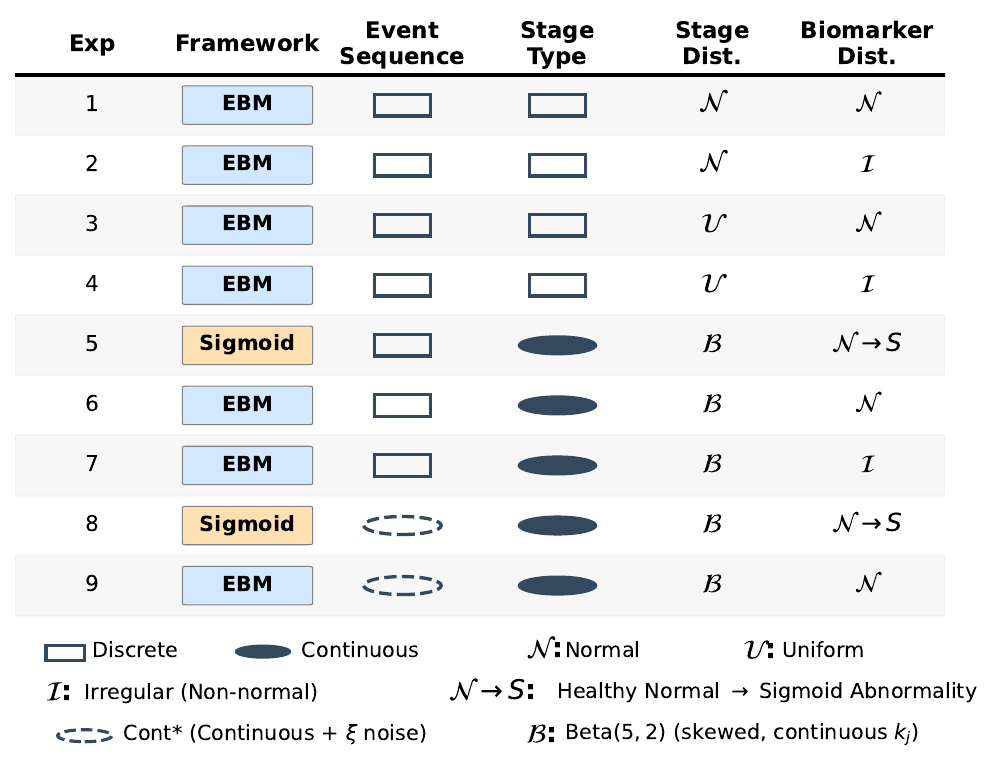} 
    \caption{\textbf{Design Matrix of Synthetic Experiments.} The nine experiments represent diverse disease hypotheses.} 
    \label{fig:exp_design_matrix}
\end{figure}

\paragraph{Benchmarking Baselines.} 
\label{sec:benchmarking baseline}
We compared \textsc{Tempo} against a comprehensive suite of EBM variants: (1) SA-EBM \citep{hao2025stageaware}; (2) UCL-KDE and UCL GMM \citep{firth2020sequences}; and (3) DEBM and DEBM GMM \citep{venkatraghavan2019disease}. For each of the nine experiments, we generated 50 independent test datasets ($50 \times 9 = 450$ total) with randomized ground-truth event sequences and patient stages to ensure statistical rigor. 

Since baseline algorithms produce only ordinal patient stage estimates, all methods, including \textsc{Tempo}, are evaluated against ordinal ground truth for the patient staging task. For experiments with continuous stages (5--9), this requires converting $k_j$ to an ordinal stage $y^*_j = |{b : \xi_b \leq k_j}|$ (the count of biomarkers whose event time has been reached). For \textsc{Tempo} training, the staging loss (Eq.~\ref{eq:stage_loss}) in Exp. 5--9 can use either the original continuous $k_j$ or the converted ordinal $y^*_j$. We tested both; training with ordinal labels yields slightly better results, likely because training and evaluation targets share the same integer scale.

\paragraph{Simulation-Based Supervised Learning (SBSL).} 
Unlike the baseline algorithms which perform inference directly on the test data, \textsc{Tempo} follows a supervised paradigm. For each experiment, we generated 1,000 additional datasets (950 for training, 50 for validation) with known ground-truth sequencing and staging labels. \textsc{Tempo} was trained for 25 iterations using an NVIDIA L40S (48 GB). The total training time for each of the nine low-dimensional experiments was 5 minutes, demonstrating significant computational efficiency. 

\paragraph{Inference and Cross-Experimental Validation.}
During the evaluation phase, we applied each of the nine trained models to all 450 test datasets, completing the full $9{\times}450$ inference in under one minute ($\approx$55 seconds). This design evaluates both in-distribution performance (e.g., Exp.~1 model tested on Exp.~1 data) and out-of-distribution generalization (e.g., Exp.~1 model tested on Exp.~9 data with continuous event times). To ensure a fair comparison, all benchmarking algorithms were evaluated on the same datasets using a dedicated CPU cluster~\citep{https://doi.org/10.21231/gnt1-hw21} with the same configurations (e.g., number of MCMC iterations) as in~\citet{hao2025stageaware}.

\subsection{Low Dimensional Results}
\label{sec:lowdim_results}

\begin{figure*}[htbp]
    \centering
    \includegraphics[width=0.9\textwidth]{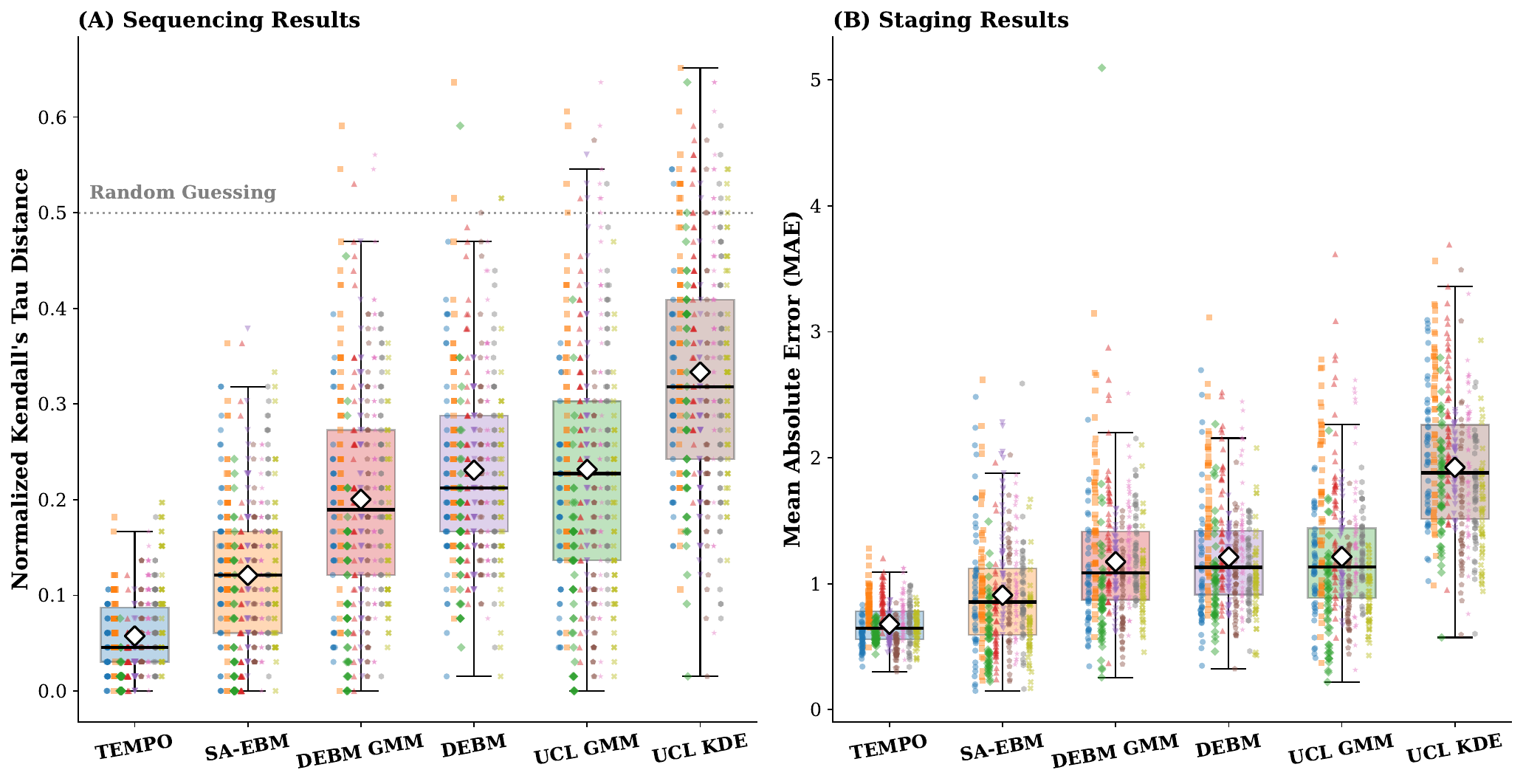} 
    \caption{\textbf{Performance on Low Dimensional Experiments.} Comparisons across (A) sequencing and (B) staging (MAE) performances. Boxes represent IQR, black lines median, and white diamonds mean. Individual results of 450 test datasets are displayed as ordered strip-plots (Exp 1 $\to$ 9). Tau = 0 indicates perfect alignment. The maximum of staging MAE is 12. \textsc{Tempo} demonstrates superior accuracy and significantly lower variance across all experimental conditions.}
    \label{fig:lowdim_results}
\end{figure*}

We report the results for the low-dimensional experiments in Fig.~\ref{fig:lowdim_results}, with per-experiment breakdowns provided in Appendix~\ref{apd:lowdim_res} (Fig.~\ref{fig:lowdim_tau_all} and~\ref{fig:lowdim_mae_all}).

\paragraph{Overall Performance.}
\textsc{Tempo} substantially outperforms all benchmarking algorithms on both the event sequencing and patient staging tasks across all nine experimental conditions. For event sequencing, \textsc{Tempo} achieved a mean normalized Kendall Tau distance of $0.057 \pm 0.004$ (mean $\pm$ 95\% CI), compared to SA-EBM at $0.121 \pm 0.007$, DEBM GMM at $0.200 \pm 0.010$, DEBM at $0.231 \pm 0.009$, UCL GMM at $0.232 \pm 0.012$, and UCL KDE at $0.333 \pm 0.011$. Relative to the state-of-the-art SA-EBM, \textsc{Tempo} reduced the normalized Tau distance by 52.89\%. For patient staging, \textsc{Tempo} achieved a mean MAE of $0.678 \pm 0.016$, followed by SA-EBM ($0.908 \pm 0.041$), DEBM GMM ($1.175 \pm 0.044$), DEBM ($1.212 \pm 0.040$), UCL GMM ($1.214 \pm 0.047$), and UCL KDE ($1.924 \pm 0.050$). \textsc{Tempo} reduced the staging MAE by 25.33\% compared to SA-EBM.

\paragraph{Per-Experiment Performance.}
Table~\ref{tab:lowdim_per_exp} in Appendix~\ref{apd:lowdim_res} presents the per-experiment results for event sequencing and patient staging. \textsc{Tempo} achieved the lowest Tau distance in all nine experiments. For sequencing, \textsc{Tempo}'s normalized Tau distance ranged from $0.013$ (Exp. 3) to $0.091$ (Exp. 9), consistently outperforming SA-EBM, whose values ranged from $0.069$ (Exp. 3) to $0.154$ (Exp. 8). For staging, \textsc{Tempo}'s MAE ranged from $0.518$ (Exp. 6) to $0.870$ (Exp. 4), while SA-EBM ranged from $0.598$ (Exp. 3) to $1.260$ (Exp. 5). In Experiments 3 and 4 (Uniform stage prior), SA-EBM achieved marginally lower staging MAE than \textsc{Tempo} ($0.598$ vs.\ $0.599$ and $0.742$ vs.\ $0.870$, respectively), suggesting that the uniform stage distribution is a relatively easy condition for the simpler EBM staging approach.

\paragraph{Cross-Experiment Generalization.}
To assess the transferability of learned progression logic across different generative frameworks, we performed cross-experiment evaluation where models trained on one experimental condition were tested on all nine conditions. Tables~\ref{tab:lowdim_cross_tau} and~\ref{tab:lowdim_cross_mae} in Appendix~\ref{apd:lowdim_res} present the complete cross-experiment matrices.

For event sequencing, the model trained on Exp. 8 (Sigmoid \& continuous event times) achieved the best cross-experiment generalization (row mean: $0.116$). Controlled pairwise comparisons of $\tau$ row means isolate the effect of each generative factor. \textbf{(1) EBM vs.\ Sigmoid}: EBM-trained models generalize worse than Sigmoid-trained models in both pairs (Exp.~6 vs.\ 5: $0.178$ vs.\ $0.148$; Exp.~9 vs.\ 8: $0.165$ vs.\ $0.116$), suggesting that training on Sigmoid-model data improves cross-experiment robustness. \textbf{(2) Normal vs.\ Non-Normal biomarkers}: Normal distributions yield worse generalization (Exp.~1 vs.\ 2: $0.150$ vs.\ $0.143$; Exp.~3 vs.\ 4: $0.156$ vs.\ $0.132$; Exp.~6 vs.\ 7: $0.178$ vs.\ $0.141$). \textbf{(3) Discrete vs.\ continuous event times}: Continuous event times improve generalization under Sigmoid (Exp.~5 vs.\ 8: $0.148$ vs.\ $0.116$) and marginally under EBM (Exp.~6 vs.\ 9: $0.178$ vs.\ $0.165$). We also report sequence MAE in Table~\ref{tab:sequence_mae_lowdim} and Fig.~\ref{fig:sequence_mae_lowdim} in Appendix~\ref{apd:lowdim_res}.

For patient staging, ordinal-stage models (Exp~1--4) achieved a mean row-mean MAE of $1.47$, compared to $1.22$ for continuous-stage models (Exp~5--8). Exp~9 is an outlier (row mean: $1.49$), closer to the ordinal group than the other continuous-stage models. The full cross-experiment stage MAE matrix is provided in Table~\ref{tab:lowdim_cross_mae} in Appendix~\ref{apd:lowdim_res}.

\paragraph{Variance and Stability.}
Beyond mean performance, \textsc{Tempo} demonstrated substantially lower variance across experimental conditions (Fig.~\ref{fig:lowdim_results}). For sequencing, \textsc{Tempo}'s standard deviation across the 450 test datasets was $0.041$, compared to $0.080$ for SA-EBM, $0.096$ for DEBM, $0.113$ for DEBM GMM, $0.119$ for UCL KDE, and $0.127$ for UCL GMM. For staging, \textsc{Tempo}'s standard deviation was $0.168$, compared to $0.446$ for SA-EBM, $0.433$ for DEBM, $0.478$ for DEBM GMM, $0.503$ for UCL GMM, and $0.543$ for UCL KDE. This lower variance, combined with superior mean performance, indicates that \textsc{Tempo} is both more accurate and more consistent across experimental conditions.

\subsection{High Dimensional Experiments}
\label{sec:highdim_exp}

To evaluate scalability to larger biomarker sets, we generated high-dimensional synthetic datasets with $B = 100$ biomarkers whose distribution parameters are exactly the same as reported in~\citet{hao2025bayesian}. Other configurations remained identical to the low-dimensional study (Section~\ref{sec:lowdim_exp}). We benchmarked all five baseline algorithms in the high-dimensional setting; 
average runtimes per test dataset were: DEBM ($4.9$ min), DEBM GMM ($5.0$ min), 
UCL GMM ($11.7$ min), UCL KDE ($12.4$ min), and SA-EBM ($55.1$ min). 
After a one-time training cost of 47 minutes per experiment, 
\textsc{Tempo} performs inference on each test dataset in less than a second.

\subsection{High Dimensional Results}
\label{sec:highdim_results}

We report the high-dimensional results in Fig.~\ref{fig:highdim_results}, with per-experiment breakdowns provided in Appendix~\ref{apd:highdim_res} (Fig.~\ref{fig:highdim_tau_all} and~\ref{fig:highdim_mae_all}).

\paragraph{Overall Performance.}
\textsc{Tempo} outperformed all baselines on both event sequencing and patient staging tasks. For sequencing, \textsc{Tempo} achieved a mean normalized Tau distance of $0.081 \pm 0.002$ (mean $\pm$ 95\% CI), compared to SA-EBM at $0.197 \pm 0.007$, DEBM GMM at $0.212 \pm 0.005$, DEBM at $0.245 \pm 0.004$, UCL GMM at $0.297 \pm 0.014$, and UCL KDE at $0.501 \pm 0.005$. This represents a 58.88\% reduction in Tau distance relative to SA-EBM. For staging, \textsc{Tempo} achieved a mean MAE of $3.24 \pm 0.06$, compared to SA-EBM ($8.33 \pm 0.43$), DEBM ($7.31 \pm 0.23$), DEBM GMM ($9.33 \pm 0.51$), UCL GMM ($21.93 \pm 1.34$), and UCL KDE ($44.59 \pm 0.79$). Notably, DEBM achieves lower staging MAE than SA-EBM in the high-dimensional setting ($7.31$ vs.\ $8.33$), reversing the low-dimensional ranking. Still, \textsc{Tempo} reduces staging MAE by 55.68\% relative to DEBM and 61.10\% relative to SA-EBM.

\paragraph{Per-Experiment Performance.}
Table~\ref{tab:highdim_per_exp} (and Figures~\ref{fig:highdim_tau_all}--\ref{fig:highdim_mae_all} for all baselines) in Appendix~\ref{apd:highdim_res} presents the per-experiment results. \textsc{Tempo} achieved the lowest Tau distance and staging MAE in all nine experiments. For sequencing, \textsc{Tempo}'s Tau distance ranged from $0.042$ (Exp 3) to $0.110$ (Exp 7), outperforming SA-EBM whose values ranged from $0.155$ (Exp 9) to $0.247$ (Exp 2). For staging, \textsc{Tempo}'s MAE ranged from $2.61$ (Exp 1) to $4.27$ (Exp 4), while SA-EBM ranged from $4.79$ (Exp 9) to $12.99$ (Exp 2).

\paragraph{Comparison with Low-Dimensional Results.}
Compared to the low-dimensional setting ($B=12$), \textsc{Tempo}'s sequencing Tau distance increased by 42.11\% (from $0.057$ to $0.081$), while SA-EBM's increased by 62.81\% (from $0.121$ to $0.197$). Unlike Tau distance, which is scale-free (a ratio of discordant pairs bounded in $[0,1]$), staging MAE scales with $B$ and must be normalized for cross-scale comparison. When normalized by $B$, \textsc{Tempo} improves by $-42.7\%$ (from $0.057$ to $0.032$),
DEBM by $-27.6\%$ ($0.101$ to $0.073$), and DEBM GMM by $-4.7\%$ ($0.098$ to $0.093$),
while SA-EBM worsens by $+10.0\%$ ($0.076$ to $0.083$), UCL GMM by $+116.8\%$ ($0.101$ to $0.219$),
and UCL KDE by $+178.1\%$ ($0.160$ to $0.446$).


\paragraph{Cross-Experiment Generalization.}
Tables~\ref{tab:highdim_cross_tau} and~\ref{tab:highdim_cross_mae} in Appendix~\ref{apd:highdim_res} present the complete cross-experiment matrices for the high-dimensional setting.

For sequencing, the model trained on Exp 8 again achieved the best cross-experiment generalization (row mean: $0.152$), followed by Exp 5 ($0.167$) and Exp 4 ($0.168$). Controlled pairwise comparisons reveal consistent patterns at scale. \textbf{(1) EBM vs.\ Sigmoid}: EBM-trained models again generalize worse in both pairs (Exp.~6 vs.\ 5: $0.184$ vs.\ $0.167$; Exp.~9 vs.\ 8: $0.199$ vs.\ $0.152$), consistent with the low-dimensional findings. \textbf{(2) Normal vs.\ Non-Normal biomarkers}: Non-Normal distributions improve generalization across all three pairs (Exp.~1 vs.\ 2: $0.193$ vs.\ $0.170$; Exp.~3 vs.\ 4: $0.187$ vs.\ $0.168$; Exp.~6 vs.\ 7: $0.184$ vs.\ $0.172$). \textbf{(3) Discrete vs.\ continuous event times}: Continuous event times improve generalization under Sigmoid (Exp.~5 vs.\ 8: $0.167$ vs.\ $0.152$), but not under EBM (Exp.~6 vs.\ 9: $0.184$ vs.\ $0.199$). We also report sequence MAE in Table~\ref{tab:sequence_mae_highdim} and Fig.~\ref{fig:sequence_mae_highdim} in Appendix~\ref{apd:highdim_res}.

For patient staging, ordinal-stage models (Exp~1--4) and continuous-stage models (Exp~5--8) achieve comparable mean row-mean MAE ($10.20$ vs.\ $9.97$). Exp~9 again emerges as an outlier (row mean: $14.37$). The full cross-experiment stage MAE matrix is provided in Appendix~\ref{apd:highdim_res}.

\paragraph{Variance and Stability.}
\textsc{Tempo} also exhibited substantially lower variance than all baselines in the high-dimensional setting. For sequencing, \textsc{Tempo}'s standard deviation across the 450 test datasets was $0.022$, compared to $0.072$ for SA-EBM. For staging, \textsc{Tempo}'s standard deviation was $0.68$, compared to $4.68$ for SA-EBM. The lower variance indicates that \textsc{Tempo}'s performance is more consistent  even as dimensionality increases.

\section{Real-World Experiments \& Results}
\label{sec:adni}
\begin{figure*}[t]
    \centering
    \includegraphics[width=0.95\textwidth]{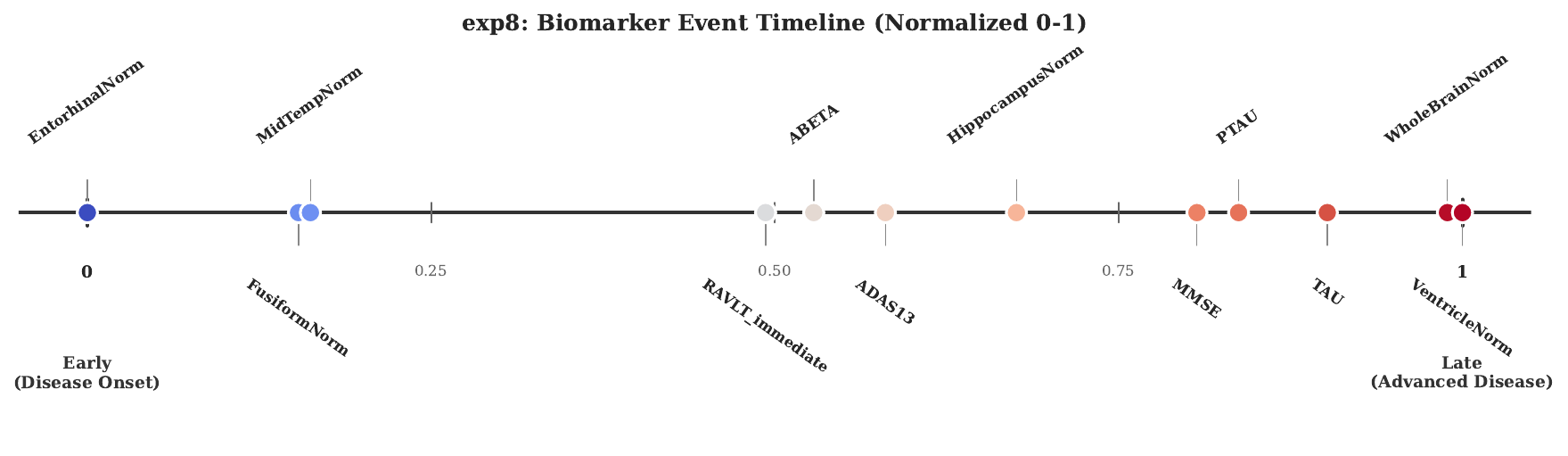}
    \caption{\textbf{Continuous Temporal Spacing of Biomarker Abnormality (Exp. 8).} The timeline illustrates the estimated time to reach a threshold of abnormality, generated by a representative model (Exp 8). The horizontal axis is min-max normalized to $[0, 1]$, where $0$ corresponds to the first biomarker becoming abnormal and $1$ to the last; larger gaps indicate greater temporal separation. Unlike discrete rankings, this visualization captures the temporal clustering of biomarker events: a dense structural onset cluster ($t < 0.2$) of medial temporal atrophy, a prolonged preclinical gap with no detected events ($t \approx 0.2$--$0.5$), a cognitive-amyloid phase ($t \approx 0.5$--$0.7$), and a terminal acceleration ($t > 0.8$) of tau pathology and global neurodegeneration.}

    \label{fig:continuous_timeline}
\end{figure*}

We used the \texttt{ADNIMERGE} table from the Alzheimer's Disease Neuroimaging Initiative \citep[ADNI,][]{mueller2005alzheimer}, last updated September 7, 2023. The ADNI was launched in 2003 as a public-private partnership, led by Principal Investigator Michael W. Weiner, MD. The primary goal of ADNI has been to test whether serial magnetic resonance imaging (MRI), positron emission tomography (PET), other biological markers, and clinical and neuropsychological assessment can be combined to measure the progression of mild cognitive impairment (MCI) and early Alzheimer’s disease (AD). 

We report our data processing procedure in Appendix~\ref{apd:adni}. To establish the characteristic sequence of AD progression, we applied all nine \textsc{Tempo} models trained on the low-dimensional synthetic experiments to the ADNI cohort. Each model produced an ordering of the 12 biomarkers. We computed biomarkers' mean ranks across models to obtain a consensus sequence (Figure~\ref{fig:adni_heatmap}) and report per-model orderings in Table~\ref{tab:biomarker_position_matrix}. See Appendix~\ref{apd:adni} for details.

\paragraph{Consensus Biomarker Ordering.}
The consensus sequence reveals a pathological cascade progressing from focal medial temporal atrophy, through cognitive decline and amyloid accumulation, to late-stage tau pathology and global neurodegeneration:

\begin{center}
\small
\textit{Entorhinal} $\to$ \textit{MidTemp} $\to$ \textit{Fusiform} $\to$ \textit{ADAS13} $\to$ \textit{RAVLT\_immediate} $\to$ \textit{ABETA} $\to$ \textit{Hippocampus} $\to$ \textit{MMSE} $\to$ \textit{PTAU} $\to$ \textit{TAU} $\to$ \textit{Ventricle} $\to$ \textit{WholeBrain}
\end{center}

The earliest events involve structural changes in the medial temporal lobe, \textit{Entorhinal} (mean rank $1.7 \pm 0.9$ SD) and \textit{MidTemp} ($2.0 \pm 1.0$), alongside cortical thinning (\textit{Fusiform}, $3.6 \pm 1.4$) and composite cognitive symptoms (\textit{ADAS13}, $3.8 \pm 1.6$). This ordering is broadly consistent with the hypothesized progression of structural MRI abnormalities from medial temporal to more lateral temporal regions~\citep{jack2010hypothetical} and that \textit{Entorhinal} gets pathological first is consistent with Braak staging~\citep{braak1991neuropathological}. 

The middle phase is characterized by episodic memory decline (\textit{RAVLT\_immediate}, $4.6 \pm 1.0$), amyloid-$\beta$ accumulation (\textit{ABETA}, $5.6 \pm 0.7$), hippocampal atrophy (\textit{Hippocampus}, $7.0 \pm 0.5$), and global cognitive decline (\textit{MMSE}, $8.0 \pm 0.5$). Since \textit{MMSE} is less sensitive to mild cognitive impairment~\citep{tombaugh1992mini}, it makes sense that its abnormality occurred the last among the three cognitive measurements. The early onset of structural MRI changes and the middle phase is consistent with the finding in \citet{jack2009serial} that neurodegeneration occurs prior to and progresses alongside cognitive decline. That \textit{ABETA} abnormality occurred prior to tau pathology is consistent with the cascade hypothesis by~\citet{jack2010hypothetical}. Note that although our result is inconsistent with the hypothetical cascade of \textit{ABETA} $\to$ CSF tau $\to$ MRI changes $\to$ cognitive decline~\citep{jack2010hypothetical}, it is broadly consistent with the ``Neurodegeneration-first biomarker model of late-onset AD''~\citep{jack2013biomarker}.

In the later phase, \textit{PTAU} ($9.1 \pm 0.6$) and \textit{TAU} ($9.8 \pm 0.4$) are inferred to become pathological after hippocampal atrophy and global cognitive decline, followed by late clustering of ventricular enlargement ($11.2 \pm 0.4$)  and whole-brain atrophy ($11.8 \pm 0.4$). While the relative placement of CSF tau biomarkers should be interpreted cautiously, this late clustering of global neurodegenerative markers is broadly consistent with evidence that ventricular enlargement increases with clinical severity and is associated with worsening cognition~\citep{jack2009serial}.

\paragraph{Continuous Temporal Spacing.}
A key advantage of \textsc{Tempo} is its ability to estimate continuous temporal spacing between biomarker events from cross-sectional data, a capability that remains challenging for traditional EBMs. Models trained on Exp 8 and 9 (continuous event times) produce such estimates. Fig.~\ref{fig:continuous_timeline} shows the Exp 8 timeline (Sigmoid measurement model); the Exp 9 timeline (EBM measurement model) is shown in Fig.~\ref{fig:continuous_timeline_exp9} (Appendix~\ref{apd:adni}).

The timeline shows that biomarker events cluster into distinct phases: (1) a structural onset cluster ($t < 0.2$) comprising medial temporal atrophy (Entorhinal, MidTemp, Fusiform); (2) a prolonged preclinical gap ($t \approx 0.2$--$0.5$) with no detected events; (3) a cognitive-amyloid phase ($t \approx 0.5$--$0.7$) encompassing episodic memory decline (RAVLT, $t= 0.49$), amyloid accumulation (ABETA, $t= 0.53$), composite cognitive symptoms (ADAS13, $t= 0.58$), and hippocampal atrophy ($t= 0.68$); and (4) a terminal acceleration ($t > 0.8$) of global cognitive decline (MMSE), tau pathology (PTAU, TAU), and global neurodegeneration (WholeBrain, Ventricle). This timeline is broadly consistent with the consensus sequence while providing temporal spacing as a direct model output rather than an artifact of rank averaging. The Exp~9 timeline (Fig.~\ref{fig:continuous_timeline_exp9}, Appendix), under the EBM measurement model, is similar to that of Exp~8.


\paragraph{Staging Validation by Diagnosis Group.}
We report the average predicted ordinal stage (out of 12) for each ADNI diagnosis group across all nine models (Table~\ref{tab:staging_by_dx}, Appendix~\ref{apd:adni}). The mean predicted stages (CN: $0.06$, EMCI: $5.52$, LMCI: $8.52$, AD: $10.68$) increase monotonically with clinical severity, demonstrating that \textsc{Tempo}'s staging aligns well with established disease categories without using diagnosis labels as an explicit training objective.

\section{Discussion}
\label{sec:discussion}
In this work, we introduced \textsc{Tempo}, a Transformer architecture for learning continuous disease progression over biomarkers from cross-sectional data. By treating biomarkers and patients as tokens and leveraging simulation-based supervised learning, \textsc{Tempo} achieves state-of-the-art performance on both event sequencing and patient staging tasks, reducing normalized Kendall Tau distance by 52.89\% and staging MAE by 25.33\% compared to SA-EBM in low-dimensional experiments, with larger reductions in high-dimensional settings (58.88\% and 61.10\%, respectively). Beyond performance gains, \textsc{Tempo} enables continuous temporal spacing estimation from purely cross-sectional data and provides a framework for rapid validation of generative hypotheses, both capabilities challenging through traditional EBMs.

A key finding from the cross-experiment evaluation is that training on more general biomarker distributions improves cross-experiment generalization for sequencing. Two factors consistently improve this generalization: the Sigmoid measurement model (vs.\ EBM binary-switch) and non-normal biomarker distributions (vs.\ Normal). The Sigmoid effect ranges from $0.017$--$0.049$ per pairwise comparison across scales (Exp.~6 vs.\ 5; Exp.~9 vs.\ 8); the non-normal effect from $0.007$--$0.037$ (Exp.~1 vs.\ 2; Exp.~3 vs.\ 4; Exp.~6 vs.\ 7). Continuous event times provide an additional benefit under the Sigmoid model (Exp.~5 vs.\ 8: $0.148\to 0.116$ at 12 biomarkers, $0.167\to 0.152$ at 100 biomarkers) but not consistently under EBM.

The mechanistic explanation unifies both effects. EBM assumes clean bimodal Gaussian distributions for pre- and post-event biomarker measurements. The Sigmoid model produces gradual post-event transitions that deviate from this assumption; non-normal distributions deviate from it in a different way. Both represent distributional diversity beyond the clean Gaussian bimodal pattern. A model trained on these richer conditions has seen more diverse measurement patterns and generalizes to the simpler EBM/Normal case. The reverse does not hold: models trained on Normal distributions (Exp.~1 and 3) achieve $\tau = 0.18$--$0.26$ when tested on non-normal data (Exp.~2 and 4), whereas models trained on non-normal distributions achieve only $\tau = 0.02$--$0.10$ on Normal data---a two- to threefold gap.


\textsc{Tempo} also scales favorably in patient staging. As shown in Section~\ref{sec:highdim_results}, \textsc{Tempo}'s normalized staging MAE decreases with dimensionality, while SA-EBM's worsens. SA-EBM jointly infers the event sequence, stage assignments, and biomarker distribution parameters via MCMC over a $B$-dimensional stage space; as $B$ grows, this joint inference becomes increasingly costly to converge. DEBM estimates biomarker distribution parameters independently as a preprocessing step and finds a consensus ordering via greedy adjacent swaps from per-patient orderings, and thus scales more favorably. This is consistent with DEBM surpassing SA-EBM in absolute staging MAE at $B=100$ ($7.31$ vs.\ $8.33$), a reversal of the low-dimensional ranking. \textsc{Tempo}'s substantially stronger relative gain ($42.7\%$ vs.\ $27.6\%$ and $4.7\%$ for DEBM and DEBM GMM) suggests that its patient-level attention benefits from additional biomarker tokens.

An important contribution of \textsc{Tempo} is its utility as a learned proxy for likelihood functions: trained on synthetic data from any generative framework, it can be applied to real-world data without requiring manual derivation of inference algorithms. This is especially valuable for frameworks such as Experiments~5--9 that introduce continuous patient stages or Sigmoid measurement models, where deriving dedicated inference algorithms would require substantial additional effort.
By training multiple \textsc{Tempo} models on different generative hypotheses and comparing their outputs on real data, researchers can empirically assess which underlying assumptions best align with observed clinical reality. 

Applied to ADNI, the nine \textsc{Tempo} models produced similar biomarker orderings (Table~\ref{tab:biomarker_position_matrix}, Appendix~\ref{apd:adni}). The consensus sequence (Section~\ref{sec:adni}) and the continuous timelines from Exp~8 (Fig.~\ref{fig:continuous_timeline}) and Exp~9 (Fig.~\ref{fig:continuous_timeline_exp9}) are broadly consistent with the ``Neurodegeneration-first biomarker model of late-onset AD''~\citep{jack2013biomarker} and with evidence that ventricular enlargement increases with clinical severity and cognitive decline~\citep{jack2009serial}. Specifically, our results suggest early medial temporal atrophy, followed by \textit{ABETA} abnormality and cognitive decline, with CSF tau pathology, ventricular expansion and global neurodegeneration clustering late in the inferred disease course.  We caution that this cohort-derived ordering should be interpreted as an empirical disease progression in ADNI rather than a definitive biological timeline of Alzheimer’s disease.



We emphasize that \textsc{Tempo} is not intended to replace EBMs or other generative frameworks. Rather, it depends on them to generate synthetic data with ground-truth labels for supervised training. The value of \textsc{Tempo} lies in its ability to perform fast, accurate inference once trained, and to enable rapid comparison across multiple generative hypotheses without requiring manual derivation of inference algorithms for each. As such, a potential criticism of \textsc{Tempo} is its dependence on the correctness of the generative model: if the generative framework is flawed, the trained model may be unreliable. However, this criticism highlights \textsc{Tempo}'s strength as a validation tool: by comparing model performance across frameworks and assessing biological plausibility on real data, researchers can identify which generative assumptions are most appropriate. Our results support this utility: models incorporating Sigmoid measurement models or non-normal biomarker distributions generalize better across conditions (Section~\ref{sec:synthetic}), suggesting these as preferred training choices when the true generative process is unknown. \textsc{Tempo} is intended to be retrained for each target dataset: biomarker distribution parameters are first estimated from the real data, then used to generate synthetic training data with ground-truth labels. Training is computationally efficient (5 minutes per experiment in the low-dimensional setting; approximately 47 minutes in the high-dimensional setting).

\section{Limitations \& Future Work}
\label{sec:limitation}
An important interpretive nuance concerns diagnostic labels. \textsc{Tempo} and SA-EBM incorporate diagnosis labels as input features during training and inference, whereas DEBM and UCL variants use them only for estimating biomarker distribution parameters. \textsc{Tempo}'s use of labels is consistent with clinical practice; future work could explore a label-free variant for settings where diagnosis is uncertain.

\textsc{Tempo} models a single population-level trajectory from cross-sectional data, accommodating neither patient subtypes nor longitudinal observations. Extending it to longitudinal data could be achieved by treating visits as tokens within patient trajectories or by anchoring timeline estimates using repeated measurements, similar to TEBM~\citep{wijeratne2023temporal}. Handling heterogeneous subtypes is a more challenging direction we leave to future work.


We validated \textsc{Tempo} on ADNI, a well-characterized Alzheimer's disease cohort. Evaluation on additional datasets, e.g., NACC~\citep{nacc_uds} and other neurodegenerative diseases (e.g., Parkinson's, Huntington's) would strengthen claims about generalizability. Our synthetic evaluation covered $B=12$ and $B=100$ biomarkers; whether \textsc{Tempo}'s advantages persist under extreme high-dimensional settings remains an open question for future work.


\acks{
We thank the CHTC at the University of Wisconsin-
Madison, the Henkaku Center and Dr. Grisha Szep for computing support. JLA was funded
by the Japan Probabilistic Computing Consortium
Association (JPCCA). 

Data collection and sharing for this project was funded by the Alzheimer's Disease Neuroimaging Initiative
(ADNI) (National Institutes of Health Grant U01 AG024904) and DOD ADNI (Department of Defense award
number W81XWH-12-2-0012). ADNI is funded by the National Institute on Aging, the National Institute of
Biomedical Imaging and Bioengineering, and through generous contributions from the following: AbbVie,
Alzheimer’s Association; Alzheimer’s Drug Discovery Foundation; Araclon Biotech; BioClinica, Inc.; Biogen;
Bristol-Myers Squibb Company; CereSpir, Inc.; Cogstate; Eisai Inc.; Elan Pharmaceuticals, Inc.; Eli Lilly and
Company; EuroImmun; F. Hoffmann-La Roche Ltd and its affiliated company Genentech, Inc.; Fujirebio; GE
Healthcare; IXICO Ltd.; Janssen Alzheimer Immunotherapy Research \& Development, LLC.; Johnson \&
Johnson Pharmaceutical Research \& Development LLC.; Lumosity; Lundbeck; Merck \& Co., Inc.; Meso
Scale Diagnostics, LLC.; NeuroRx Research; Neurotrack Technologies; Novartis Pharmaceuticals
Corporation; Pfizer Inc.; Piramal Imaging; Servier; Takeda Pharmaceutical Company; and Transition
Therapeutics. The Canadian Institutes of Health Research is providing funds to support ADNI clinical sites
in Canada. Private sector contributions are facilitated by the Foundation for the National Institutes of Health
(www.fnih.org). The grantee organization is the Northern California Institute for Research and Education,
and the study is coordinated by the Alzheimer’s Therapeutic Research Institute at the University of Southern
California. ADNI data are disseminated by the Laboratory for Neuro Imaging at the University of Southern
California.}

\bibliography{chil-sample}

\onecolumn
\appendix
\clearpage
\section{\textsc{TEMPO Architecture}}
\label{app:tempo_arch}
\begin{figure*}[htbp]
    \centering
    \begin{tikzpicture}[scale=1.0, every node/.style={scale=1.0},
        font=\scriptsize,
        box/.style={draw, rounded corners, align=center, inner sep=3pt, minimum width=1.6cm, minimum height=0.45cm, fill=white},
        attnbox/.style={draw, rounded corners, align=center, inner sep=3pt, minimum width=1.6cm, minimum height=0.45cm, fill=orange!15},
        ffnbox/.style={draw, rounded corners, align=center, inner sep=3pt, minimum width=1.6cm, minimum height=0.45cm, fill=yellow!18},
        normbox/.style={draw, rounded corners, align=center, inner sep=3pt, minimum width=1.6cm, minimum height=0.45cm, fill=red!10},
        stagebox/.style={draw, rounded corners, align=center, inner sep=3pt, minimum width=1.6cm, minimum height=0.45cm, fill=violet!12},
        outputbox/.style={draw, rounded corners, align=center, inner sep=3pt, minimum width=1.6cm, minimum height=0.45cm, fill=green!15},
        graybox/.style={draw, rounded corners, align=center, inner sep=2pt, minimum width=1.2cm, minimum height=0.35cm, fill=gray!12},
        arrow/.style={-Latex, semithick},
        dashedarrow/.style={-Latex, semithick, densely dashed, purple!70!black},
        lbl/.style={font=\tiny},
        mathlbl/.style={font=\tiny, text=black!80}
        ]
        
        \fill[blue!5, rounded corners=5pt] (-1.5,-0.6) rectangle (1.8,7.8);
        \node[font=\small\bfseries] (title_a) at (0.15,7.4) {Event Sequencing};
        
        \node[box] (input_r) at (0.15,0) {Input Embed};
        \node[mathlbl, below=0.08cm of input_r] {$\mathbf{X} \in \mathbb{R}^{J \times (B+1)}$};
        
        \node[box, above=0.4cm of input_r] (enc_r) {Patient Encoder};
        \node[graybox, above=0.3cm of enc_r] (pool_r) {Mean Pool};
        \node[mathlbl, left=0.1cm of pool_r] {$\text{avg}_J$};
        
        \node[circle, draw, inner sep=1pt, font=\tiny, fill=white] (oplus_r) at ($(pool_r.north)+(0,0.3)$) {$+$};
        \node[lbl, left=0.15cm of oplus_r] {Pos};
        
        \node[attnbox, above=0.35cm of oplus_r] (attn_r) {Biomarker\\Transformer};
        \node[normbox, above=0.2cm of attn_r] (norm1_r) {Add \& Norm};
        \node[ffnbox, above=0.2cm of norm1_r] (ffn_r) {FFN};
        \node[normbox, above=0.2cm of ffn_r] (norm2_r) {Add \& Norm};
        
        \node[outputbox, above=0.6cm of norm2_r] (out_r) {Ranking Head};
        \node[mathlbl, above=0.05cm of out_r] {$\mathbf{s} \in \mathbb{R}^B$};
        
        \draw[arrow] (input_r) -- (enc_r);
        \draw[arrow] (enc_r) -- (pool_r);
        \draw[arrow] (pool_r) -- (oplus_r);
        \draw[arrow] (oplus_r) -- (attn_r);
        \draw[arrow] (attn_r) -- (norm1_r);
        \draw[arrow] (norm1_r) -- (ffn_r);
        \draw[arrow] (ffn_r) -- (norm2_r);
        \draw[arrow] (norm2_r) -- (out_r);
        \node[font=\scriptsize\bfseries] at (0.15,-0.9) {(a)};
        
        \fill[violet!5, rounded corners=5pt] (2.5,-0.6) rectangle (7.8,7.8);
        \node[font=\small\bfseries] at (5.15,7.4) {Patient Staging};
        
        \node[box] (input_s) at (5.15,0) {Input Embed};
        \node[mathlbl, below=0.08cm of input_s] {$\mathbf{X} \in \mathbb{R}^{J \times (B+1)}$};

        \node[stagebox, above=0.6cm of input_s, fill=purple!10, minimum width=3cm] (abnorm) {Abnormality Detector $\sigma$};
        \node[mathlbl, right=0.02cm of abnorm, align=left] {$\sigma(x_{j,b}, d_j, \sigma(s_b)) \in [0,1]$};
        
        \node[box, above=0.4cm of abnorm] (stage_enc) {Stage Encoder};

        \node[attnbox, above=0.4cm of stage_enc] (attn_s) {Stage Transformer\\(Patient Attention)};
        \node[normbox, above=0.25cm of attn_s] (norm1_s) {Add \& Norm};
        \node[ffnbox, above=0.25cm of norm1_s] (ffn_s) {FFN};
        \node[normbox, above=0.25cm of ffn_s] (norm2_s) {Add \& Norm};
        
        \node[outputbox, above=0.45cm of norm2_s] (out_s) {Stage Head};
        \node[mathlbl, above=0.08cm of out_s] {$\hat{\mathbf{y}} \in \mathbb{R}^J$};
        
        \draw[arrow] (input_s) -- (abnorm);
        \draw[arrow] (abnorm) -- (stage_enc);
        \draw[arrow] (stage_enc) -- (attn_s);
        \draw[arrow] (attn_s) -- (norm1_s);
        \draw[arrow] (norm1_s) -- (ffn_s);
        \draw[arrow] (ffn_s) -- (norm2_s);
        \draw[arrow] (norm2_s) -- (out_s);
        \node[font=\scriptsize\bfseries] at (5.15,-0.9) {(b)};
        
        \draw[dashedarrow] (out_r.east) -- ++(0.9,0.0) |- (abnorm.west)
            node[pos=0.7, left, lbl, text=purple!70!black, xshift=-0.2cm, yshift=0.5cm] {$\mathbf{s}$};
        
    \end{tikzpicture}
    \caption{\textbf{\textsc{Tempo} Architecture.}
(a)~\textbf{Sequencing Branch:} Patient data is encoded and mean-pooled across the cohort to produce per-biomarker tokens; a Transformer processes these tokens to output ranking scores $\mathbf{s} \in \mathbb{R}^B$ encoding the temporal order of biomarker abnormality.
(b)~\textbf{Staging Branch:} A shared \textit{Abnormality Detector} takes the raw measurement $x_{j,b}$, disease label $d_j$, and sigmoid-normalized ranking score $\sigma(s_b)$ to produce per-biomarker abnormality probabilities. A \textit{Stage Encoder} projects each patient's abnormality profile into a latent space, which is then refined by a \textit{Stage Transformer} via cross-patient attention to predict individual stages.
\textbf{Notations}: $B$: number of biomarkers; $J$: participants per batch; $B+1$ includes the binary disease label. For implementation, please refer to \url{https://github.com/jpcca/tempo/blob/main/tempo.py}.}
    \label{fig:architecture}
\end{figure*}
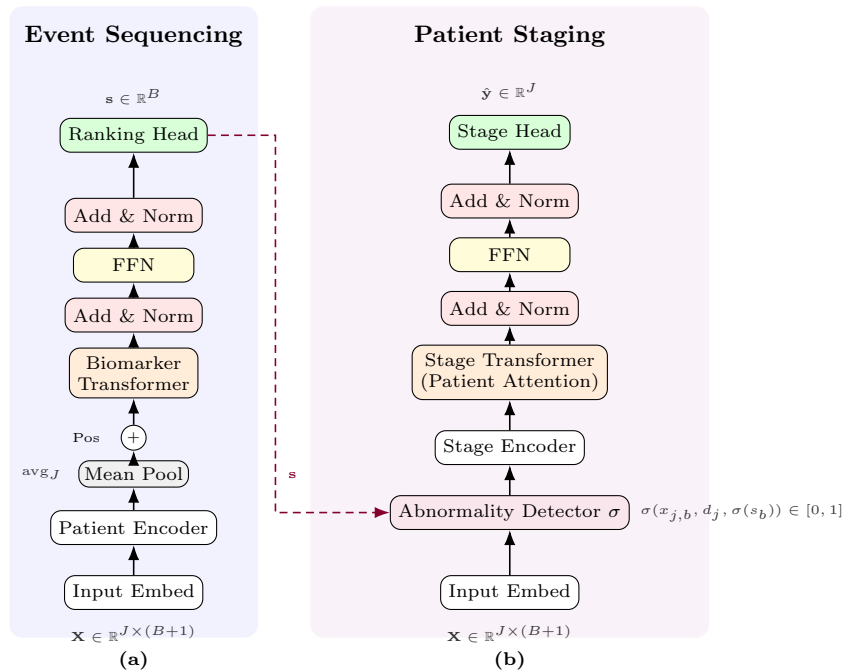

\section{Experiment Setup}
\label{app:experiment_setup}

Table~\ref{tab:synthetic_configs} lists the nine synthetic experiments and their key generative dimensions. Throughout, we use \emph{pre-event} ($\phi_b$) and \emph{post-event} ($\theta_b$) to describe the state of a biomarker, and \emph{healthy} vs.\ \emph{diseased} for participant status.

\begin{table}[htbp]
\centering
\caption{Configuration of synthetic experiments. Experiments~1 \& 3 follow standard EBM assumptions~\citep{fonteijn2012event}: discrete event times, ordinal stages with either a Dirichlet-Multinomial (DM) or Uniform stage prior, and Gaussian biomarker distributions. Experiments~2 \& 4 introduce a misspecification: non-Gaussian biomarker distributions. Experiments~5--9 introduce \textbf{continuous patient stages} ($k_j\!\sim\!\mathrm{Beta}(5,2)\cdot B$), which no baseline algorithm models explicitly. Within this group, Experiments~5 and~8 apply a Sigmoid measurement model (full misspecification relative to EBM), while Experiments~6, 7, and~9 retain the EBM binary-switch measurement with continuous stages (partial misspecification). Experiments~8 and~9 further use near-normally distributed continuous event times. $^\dagger$Pre-event baseline $\mathcal{N}(\phi_b^\mu,\phi_b^\sigma)$ for all participants; diseased participants additionally receive a sigmoid deviation (see Sigmoid measurement model).}
\label{tab:synthetic_configs}
\small
\begin{tabular}{@{}clllll@{}}
\toprule
\textbf{Exp.} & \textbf{Event Times $\xi_b$} & \textbf{Stage Type} & \textbf{Stage Prior} & \textbf{Meas.\ Model} & \textbf{Biomarker Dist.} \\
\midrule
1 & Discrete $\{1,\ldots,B\}$ & Ordinal    & Dir.-Mult.          & EBM     & Normal \\
2 & Discrete $\{1,\ldots,B\}$ & Ordinal    & Dir.-Mult.          & EBM     & Non-Normal \\
3 & Discrete $\{1,\ldots,B\}$ & Ordinal    & Uniform             & EBM     & Normal \\
4 & Discrete $\{1,\ldots,B\}$ & Ordinal    & Uniform             & EBM     & Non-Normal \\
5 & Discrete $\{1,\ldots,B\}$ & Continuous & Beta$(5,2)$         & Sigmoid & Normal $\to$ Sigmoid$^\dagger$ \\
6 & Discrete $\{1,\ldots,B\}$ & Continuous & Beta$(5,2)$         & EBM     & Normal \\
7 & Discrete $\{1,\ldots,B\}$ & Continuous & Beta$(5,2)$         & EBM     & Non-Normal \\
8 & Beta$(2,2)\!\cdot\!B$ + noise & Continuous & Beta$(5,2)$     & Sigmoid & Normal $\to$ Sigmoid$^\dagger$ \\
9 & Beta$(2,2)\!\cdot\!B$ + noise & Continuous & Beta$(5,2)$     & EBM     & Normal \\
\bottomrule
\end{tabular}
\end{table}

\paragraph{Event times $\xi_b$.}
For discrete experiments (1--7), the event time of biomarker~$b$ equals its ordinal rank in the ground-truth sequence: $\xi_b \in \{1,\ldots,B\}$. 

For continuous experiments (8--9), event times are drawn from a near-normal distribution: $\xi_b \sim \mathrm{Beta}(2,2)\cdot B$, giving values distributed roughly symmetrically over $[0,B]$. An independent per-participant noise term $\epsilon_b \sim \mathcal{N}(0,\,B\cdot 0.05)$ is then added to each $\xi_b$ in these two experiments.

\paragraph{Patient disease stage $k_j$.}
For ordinal experiments (1--4), each diseased participant is assigned $k_j \in \{1,\ldots,B\}$:
\begin{itemize}
  \item \emph{Dirichlet-Multinomial (Dir.-Mult.) prior} (exp.\ 1, 2): 
stage probabilities $\boldsymbol{\pi}$ are drawn from 
$\mathrm{Dirichlet}(\boldsymbol{\alpha})$, where the concentration vector 
$\boldsymbol{\alpha}$ is constructed to follow a Gaussian-shaped profile 
over stages, with $\alpha_k \in [0.35,\,4.25]$, centered at $(B-1)/2$, and 
width $\sigma_\alpha = B/6$. This induces a roughly bell-shaped 
stage distribution.
  \item \emph{Uniform prior} (exp.\ 3, 4): $\boldsymbol{\pi} \sim \mathrm{Dirichlet}(100,\ldots,100)$, giving approximately equal stage frequencies.
\end{itemize}
For continuous experiments (5--9), $k_j \sim \mathrm{Beta}(5,2)\cdot B$, clipped to $(0, B]$, for diseased participants (skewed toward later stages) and $k_j = 0$ for healthy participants.

\paragraph{EBM measurement model.}
Each biomarker measurement is drawn from either the pre-event or post-event component depending on whether the biomarker's event time has been reached:
\begin{equation}
  x_{j,b} \;\sim\;
  \begin{cases}
    \mathcal{N}(\theta_b^{\mu},\,\theta_b^{\sigma}) & \text{if } k_j \geq \xi_b \quad\text{(post-event state),} \\
    \mathcal{N}(\phi_b^{\mu},\,\phi_b^{\sigma})     & \text{otherwise} \quad\text{(pre-event state).}
  \end{cases}
\end{equation}
For non-normal experiments (2, 4, 7), the Gaussian is replaced by a randomly selected composite distribution. At the start of each dataset, each biomarker independently and uniformly draws one of six distribution families listed below; all families use the biomarker's own $(\mu, \sigma)$ as location and scale. After sampling, additive noise $\mathcal{N}(0,(0.2\sigma)^2)$ is applied and the result is clipped to $[\mu\pm5\sigma]$. Here $s \in \{-1,+1\}$ and $v \in \{0,2\sigma\}$ are drawn uniformly per element.

\begin{enumerate}[leftmargin=*, itemsep=2pt]
  \item \textbf{Triangular / Normal / Exponential} (equal thirds):
    Tri$(\mu{-}2\sigma,\,\mu{-}1.5\sigma,\,\mu)$;\;
    $\mathcal{N}(\mu{+}\sigma,\,(0.3\sigma)^2)$;\;
    Exp$(0.7\sigma)+(\mu{-}0.5\sigma)$.

  \item \textbf{Pareto / Uniform / Logistic} (equal thirds):
    Pareto$(1.5)\cdot\sigma+(\mu{-}2\sigma)$;\;
    $\mathcal{U}(\mu{-}1.5\sigma,\,\mu{+}1.5\sigma)$;\;
    Logistic$(\mu,\,\sigma)$.

  \item \textbf{Beta / Signed-Exp / Spike-Normal} (equal thirds):
    Beta$(0.5,0.5)\cdot4\sigma+(\mu{-}2\sigma)$;\;
    Exp$(0.4\sigma)\cdot s+\mu$;\;
    $\mathcal{N}(\mu,\,(0.5\sigma)^2)+v$.

  \item \textbf{Gamma / Weibull / Shifted-Normal} (equal thirds):
    Gamma$(2,\,0.5\sigma)+(\mu{-}\sigma)$;\;
    Weibull$(1)\cdot\sigma+(\mu{-}\sigma)$;\;
    $\mathcal{N}(\mu,\,(0.5\sigma)^2)\pm\sigma$.

  \item \textbf{Heavy-tailed Cauchy} (single component):
    Cauchy$(\mu,\sigma)+\mathcal{N}(0,(0.2\sigma)^2)$, clipped to $[\mu\pm4\sigma]$.

  \item \textbf{Bimodal} (10\% / 90\% split):
    $\mathcal{N}(\mu,\,(0.2\sigma)^2)$ [10\%]\;+\;
    Logistic$(\mu{+}\sigma,\,2\sigma)$ [90\%].
\end{enumerate}

\paragraph{Sigmoid measurement model.}
All participants receive a pre-event baseline: $x_{j,b}^{0} \sim \mathcal{N}(\phi_b^{\mu},\phi_b^{\sigma})$. Healthy participants retain this value ($x_{j,b} = x_{j,b}^{0}$). For diseased participants, a sigmoid shift modulates the deviation from the pre-event state:
\begin{equation}
  x_{j,b} \;=\; x_{j,b}^{0} \;+\; \delta_b \cdot \frac{R_b}{1 + e^{-\rho_b(k_j - \xi_b)}},
\end{equation}
where $R_b = \theta_b^{\mu} - \phi_b^{\mu}$ is the maximum post-event shift, $\rho_b = \max\!\bigl(1,\,|R_b|\,/\sqrt{(\theta_b^{\sigma})^2+(\phi_b^{\sigma})^2}\bigr)$ controls the transition rate, and $\delta_b = (-1)^{Z_b}$ with $Z_b \sim \mathrm{Bernoulli}(0.5)$, drawn independently per biomarker once per dataset. When $k_j \ll \xi_b$, the shift approaches~0 (pre-event state); when $k_j \gg \xi_b$, it approaches $\delta_b R_b$ (post-event state). This model is inspired by~\citet{jack2010hypothetical}, \citet{venkatraghavan2019disease}, and \citet{young2015simulation}.

\paragraph{Biomarker parameters.}

Pre-event ($\phi_b$) and post-event ($\theta_b$) parameters for the 12 low-dimensional biomarkers are estimated from ADNI using the standard EBM algorithm (Section~\ref{apd:adni}). For the 100-biomarker high-dimensional experiments, parameters are sampled as described in~\citet{hao2025bayesian}.

\clearpage
\section{Experimental Results}
\label{apd:res}

\subsection{Low Dimensional}
\label{apd:lowdim_res}

\begin{figure*}[htbp]
    \centering
    \includegraphics[width=1\textwidth]{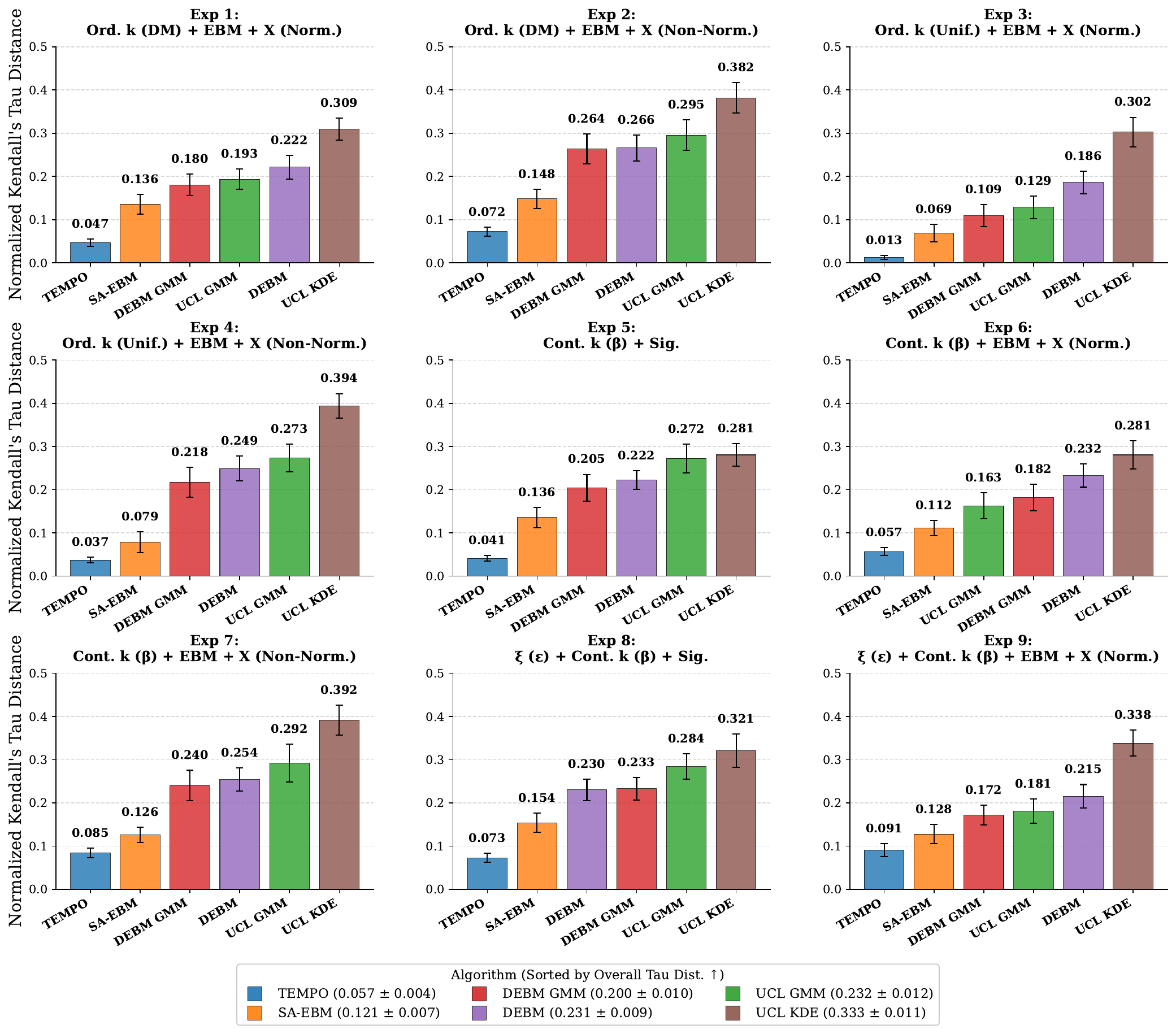} 
    \caption{\textbf{Detailed Event Sequencing Performance (Low-Dimensional).} Comparison of normalized Tau distance (lower is better) across all nine synthetic experiments ($B=12$). Each panel corresponds to a unique generative hypothesis defined in Table \ref{tab:synthetic_configs}, ranging from standard EBM assumptions (Exp 1--4) to continuous-stage frameworks with Sigmoid and EBM measurement models (Exp 5--9). \textsc{Tempo} (blue) consistently achieves the lowest distance across all conditions, demonstrating superior robustness even when the underlying biomarker distributions are non-Gaussian or non-linear. Error bars represent 95\% confidence intervals across the 50 test datasets in each experiment.}
    \label{fig:lowdim_tau_all}
\end{figure*}

\begin{figure*}[htbp]
    \centering
    \includegraphics[width=1\textwidth]{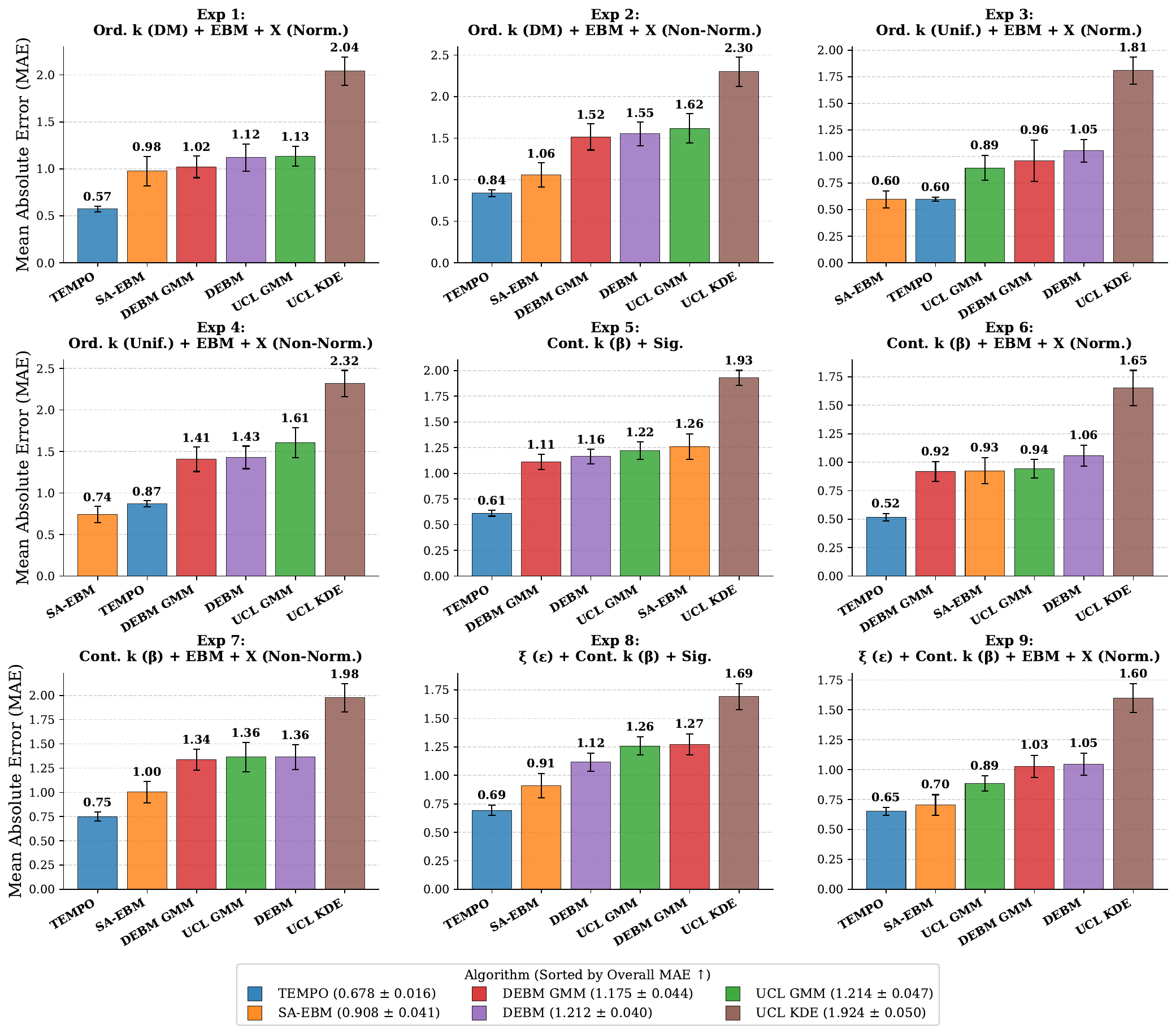} 
    \caption{\textbf{Detailed Patient Staging Performance (Low-Dimensional).} Comparison of staging Mean Absolute Error (MAE, lower is better) across nine experimental conditions ($B=12$). The grid illustrates \textsc{Tempo}'s performance stability across varying patient stage distributions and biomarker transition logics. \textsc{Tempo} maintains a significant performance margin over baseline EBMs in continuous-stage experiments (Exp 5--9), which introduce varying degrees of model misspecification. Error bars represent 95\% confidence intervals across the 50 test datasets in each experiment.}
    \label{fig:lowdim_mae_all}
\end{figure*}


\begin{table*}[htbp]
\centering
\caption{\textbf{Per-Experiment Performance (Low Dimensional).} Mean normalized Tau distance and staging MAE with 95\% confidence intervals. Since we have 12 biomarkers, the maximum stage error is 12. Lower values indicate better performance. \textsc{Tempo} achieves the lowest Tau distance across all experiments. For staging MAE, SA-EBM achieves marginally lower error in Experiments~3 and~4 (Uniform stage prior).}
\label{tab:lowdim_per_exp}
\small
\begin{tabular}{@{}lcccc@{}}
\toprule
& \multicolumn{2}{c}{\textbf{Tau Distance} $\downarrow$} & \multicolumn{2}{c}{\textbf{Staging MAE} $\downarrow$} \\
\cmidrule(lr){2-3} \cmidrule(lr){4-5}
\textbf{Exp} & \textsc{Tempo} & SA-EBM & \textsc{Tempo} & SA-EBM \\
\midrule
1 & $\mathbf{0.047 \pm 0.008}$ & $0.136 \pm 0.023$ & $\mathbf{0.573 \pm 0.029}$ & $0.976 \pm 0.156$ \\
2 & $\mathbf{0.072 \pm 0.010}$ & $0.148 \pm 0.022$ & $\mathbf{0.839 \pm 0.041}$ & $1.058 \pm 0.145$ \\
3 & $\mathbf{0.013 \pm 0.004}$ & $0.069 \pm 0.020$ & $0.599 \pm 0.019$ & $\mathbf{0.598 \pm 0.079}$ \\
4 & $\mathbf{0.037 \pm 0.006}$ & $0.079 \pm 0.024$ & $0.870 \pm 0.035$ & $\mathbf{0.742 \pm 0.097}$ \\
5 & $\mathbf{0.041 \pm 0.007}$ & $0.136 \pm 0.024$ & $\mathbf{0.612 \pm 0.028}$ & $1.260 \pm 0.126$ \\
6 & $\mathbf{0.057 \pm 0.009}$ & $0.112 \pm 0.017$ & $\mathbf{0.518 \pm 0.032}$ & $0.926 \pm 0.113$ \\
7 & $\mathbf{0.085 \pm 0.011}$ & $0.126 \pm 0.018$ & $\mathbf{0.751 \pm 0.044}$ & $1.002 \pm 0.110$ \\
8 & $\mathbf{0.073 \pm 0.010}$ & $0.154 \pm 0.022$ & $\mathbf{0.694 \pm 0.045}$ & $0.909 \pm 0.106$ \\
9 & $\mathbf{0.091 \pm 0.014}$ & $0.128 \pm 0.022$ & $\mathbf{0.652 \pm 0.032}$ & $0.704 \pm 0.087$ \\
\bottomrule
\end{tabular}
\end{table*}

\begin{table*}[t]
\centering
\caption{\textbf{Cross-Condition Sequence MAE Generalization (Low Dimensional).} Mean Absolute Error (MAE) between predicted and ground-truth event times across training and testing environments, normalized to the ground-truth event-time scale. Diagonal entries (bold) represent in-domain performance; off-diagonal entries indicate out-of-distribution generalization.}
\label{tab:sequence_mae_lowdim}
\resizebox{\textwidth}{!}{
\begin{tabular}{lcccccccccc}
\toprule
\textbf{Trained \textbackslash Tested} & \textbf{Exp 1} & \textbf{Exp 2} & \textbf{Exp 3} & \textbf{Exp 4} & \textbf{Exp 5} & \textbf{Exp 6} & \textbf{Exp 7} & \textbf{Exp 8} & \textbf{Exp 9} & \textbf{Row Mean} \\
\midrule
\textbf{Exp 1} & \textbf{0.693} & 1.754 & 0.475 & 1.716 & 2.171 & 1.010 & 1.812 & 1.977 & 0.961 & 1.397 \\
\textbf{Exp 2} & 0.954 & \textbf{0.994} & 0.641 & 0.774 & 2.732 & 1.288 & 1.225 & 2.589 & 1.459 & 1.406 \\
\textbf{Exp 3} & 0.951 & 1.724 & \textbf{0.397} & 1.620 & 2.138 & 1.307 & 1.781 & 1.973 & 1.361 & 1.473 \\
\textbf{Exp 4} & 1.039 & 1.049 & 0.498 & \textbf{0.629} & 2.550 & 1.421 & 1.313 & 2.447 & 1.493 & 1.382 \\
\textbf{Exp 5} & 1.492 & 2.216 & 1.622 & 2.458 & \textbf{0.639} & 1.224 & 2.155 & 0.773 & 1.102 & 1.520 \\
\textbf{Exp 6} & 1.051 & 1.831 & 1.297 & 2.055 & 2.718 & \textbf{1.009} & 1.813 & 2.357 & 1.029 & 1.685 \\
\textbf{Exp 7} & 1.057 & 1.191 & 0.703 & 1.038 & 2.623 & 1.088 & \textbf{1.106} & 2.458 & 1.392 & 1.406 \\
\textbf{Exp 8} & 1.353 & 2.013 & 1.306 & 2.242 & 0.778 & 1.088 & 1.946 & \textbf{0.672} & 0.932 & 1.370 \\
\textbf{Exp 9} & 1.196 & 1.856 & 1.456 & 2.039 & 3.110 & 0.891 & 1.840 & 2.650 & \textbf{0.828} & 1.763 \\
\midrule
\textbf{Col Mean} & 1.087 & 1.626 & 0.933 & 1.619 & 2.162 & 1.147 & 1.666 & 1.988 & 1.173 & \textbf{1.489} \\
\bottomrule
\end{tabular}
}
\end{table*}

\begin{table*}[t]
\centering
\caption{\textbf{Cross-Condition Tau Distance Generalization (Low Dimensional).} Average normalized Tau distance across training and testing environments. Diagonal entries (bold) represent in-domain performance. This matrix illustrates the capability of transferred progression logic across disparate pathological models.}
\label{tab:lowdim_cross_tau}
\resizebox{\textwidth}{!}{
\begin{tabular}{lcccccccccc}
\toprule
\textbf{Trained \textbackslash Tested} & \textbf{Exp 1} & \textbf{Exp 2} & \textbf{Exp 3} & \textbf{Exp 4} & \textbf{Exp 5} & \textbf{Exp 6} & \textbf{Exp 7} & \textbf{Exp 8} & \textbf{Exp 9} & \textbf{Row Mean} \\
\midrule
\textbf{Exp 1} & \textbf{0.047} & 0.188 & 0.028 & 0.200 & 0.242 & 0.059 & 0.194 & 0.291 & 0.102 & 0.150 \\
\textbf{Exp 2} & 0.061 & \textbf{0.072} & 0.034 & 0.058 & 0.301 & 0.087 & 0.099 & 0.376 & 0.196 & 0.143 \\
\textbf{Exp 3} & 0.065 & 0.178 & \textbf{0.013} & 0.178 & 0.227 & 0.092 & 0.195 & 0.286 & 0.169 & 0.156 \\
\textbf{Exp 4} & 0.072 & 0.079 & 0.020 & \textbf{0.037} & 0.246 & 0.100 & 0.108 & 0.341 & 0.183 & 0.132 \\
\textbf{Exp 5} & 0.112 & 0.219 & 0.150 & 0.248 & \textbf{0.041} & 0.109 & 0.224 & 0.085 & 0.149 & 0.148 \\
\textbf{Exp 6} & 0.065 & 0.188 & 0.100 & 0.212 & 0.318 & \textbf{0.057} & 0.199 & 0.352 & 0.106 & 0.178 \\
\textbf{Exp 7} & 0.064 & 0.086 & 0.043 & 0.086 & 0.282 & 0.082 & \textbf{0.085} & 0.374 & 0.172 & 0.141 \\
\textbf{Exp 8} & 0.082 & 0.178 & 0.051 & 0.182 & 0.044 & 0.095 & 0.204 & \textbf{0.073} & 0.132 & 0.116 \\
\textbf{Exp 9} & 0.051 & 0.182 & 0.026 & 0.186 & 0.327 & 0.060 & 0.192 & 0.371 & \textbf{0.091} & 0.165 \\
\midrule
\textbf{Col Mean} & 0.069 & 0.152 & 0.052 & 0.154 & 0.225 & 0.083 & 0.167 & 0.283 & 0.144 & \textbf{0.148} \\
\bottomrule
\end{tabular}
}
\end{table*}

\begin{table*}[t]
\centering
\caption{\textbf{Cross-Condition Staging MAE Generalization (Low Dimensional).} Mean Absolute Error (MAE) between predicted and ground-truth patient stages across training and testing environments. Diagonal entries (bold) represent in-domain performance. Values indicate the stability of staging accuracy under varying pathological assumptions.}
\label{tab:lowdim_cross_mae}
\resizebox{\textwidth}{!}{
\begin{tabular}{lcccccccccc}
\toprule
\textbf{Trained \textbackslash Tested} & \textbf{Exp 1} & \textbf{Exp 2} & \textbf{Exp 3} & \textbf{Exp 4} & \textbf{Exp 5} & \textbf{Exp 6} & \textbf{Exp 7} & \textbf{Exp 8} & \textbf{Exp 9} & \textbf{Row Mean} \\
\midrule
\textbf{Exp 1} & \textbf{0.573} & 1.022 & 0.671 & 1.179 & 3.051 & 0.643 & 1.116 & 3.819 & 0.842 & 1.435 \\
\textbf{Exp 2} & 0.682 & \textbf{0.839} & 0.825 & 1.063 & 2.703 & 0.947 & 1.085 & 3.506 & 1.547 & 1.466 \\
\textbf{Exp 3} & 0.840 & 1.359 & \textbf{0.599} & 1.125 & 3.123 & 0.863 & 1.413 & 3.703 & 0.999 & 1.558 \\
\textbf{Exp 4} & 0.852 & 1.059 & 0.647 & \textbf{0.870} & 2.308 & 1.003 & 1.300 & 3.162 & 1.426 & 1.403 \\
\textbf{Exp 5} & 1.322 & 1.694 & 1.329 & 1.865 & \textbf{0.612} & 0.860 & 1.094 & 1.290 & 1.344 & 1.268 \\
\textbf{Exp 6} & 1.004 & 1.296 & 1.167 & 1.502 & 1.555 & \textbf{0.518} & 0.900 & 2.412 & 1.160 & 1.279 \\
\textbf{Exp 7} & 1.120 & 1.248 & 1.309 & 1.521 & 1.185 & 0.573 & \textbf{0.751} & 1.998 & 1.255 & 1.218 \\
\textbf{Exp 8} & 0.952 & 1.713 & 1.015 & 1.823 & 0.695 & 0.882 & 1.376 & \textbf{0.694} & 0.899 & 1.117 \\
\textbf{Exp 9} & 0.642 & 1.012 & 0.722 & 1.137 & 3.483 & 0.591 & 0.994 & 4.213 & \textbf{0.652} & 1.494 \\
\midrule
\textbf{Col Mean} & 0.887 & 1.249 & 0.920 & 1.343 & 2.079 & 0.764 & 1.114 & 2.755 & 1.125 & \textbf{1.360} \\
\bottomrule
\end{tabular}
}
\end{table*}

\begin{figure*}[htbp]
    \centering
    \includegraphics[width=0.85\textwidth]{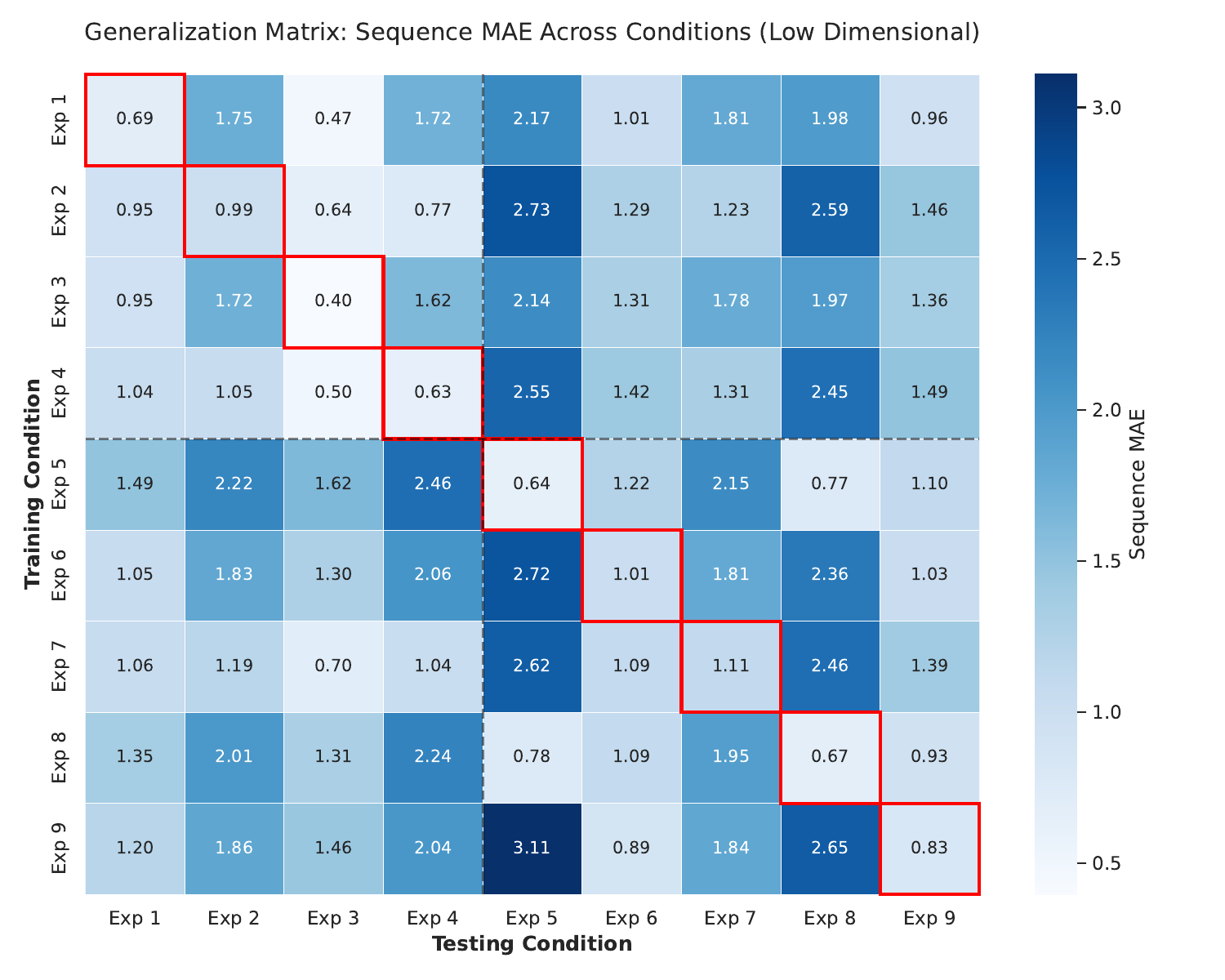} 
    \caption{\textbf{Cross-Condition Sequence Generalization (Low Dimensional).} Heatmap of Sequence MAE across 81 training and testing permutations ($B=12$). Red squares along the diagonal denote in-domain performance. Columns 5 and 8 (Sigmoid-generated test data with continuous event times) tend to have higher col-mean errors due to scale differences from discrete-event-time experiments. Among trained models, Exp~8 (Sigmoid with continuous event times) achieves the lowest row-mean error ($1.370$), while Exp~9 (EBM with continuous event times, row mean $1.763$) generalizes least well, confirming that the Sigmoid measurement model---not continuous event times alone---drives cross-experiment robustness.}
    \label{fig:sequence_mae_lowdim}
\end{figure*}

\clearpage
\subsection{High Dimensional}
\label{apd:highdim_res}

\begin{figure*}[htbp]
    \centering
    \includegraphics[width=0.9\textwidth]{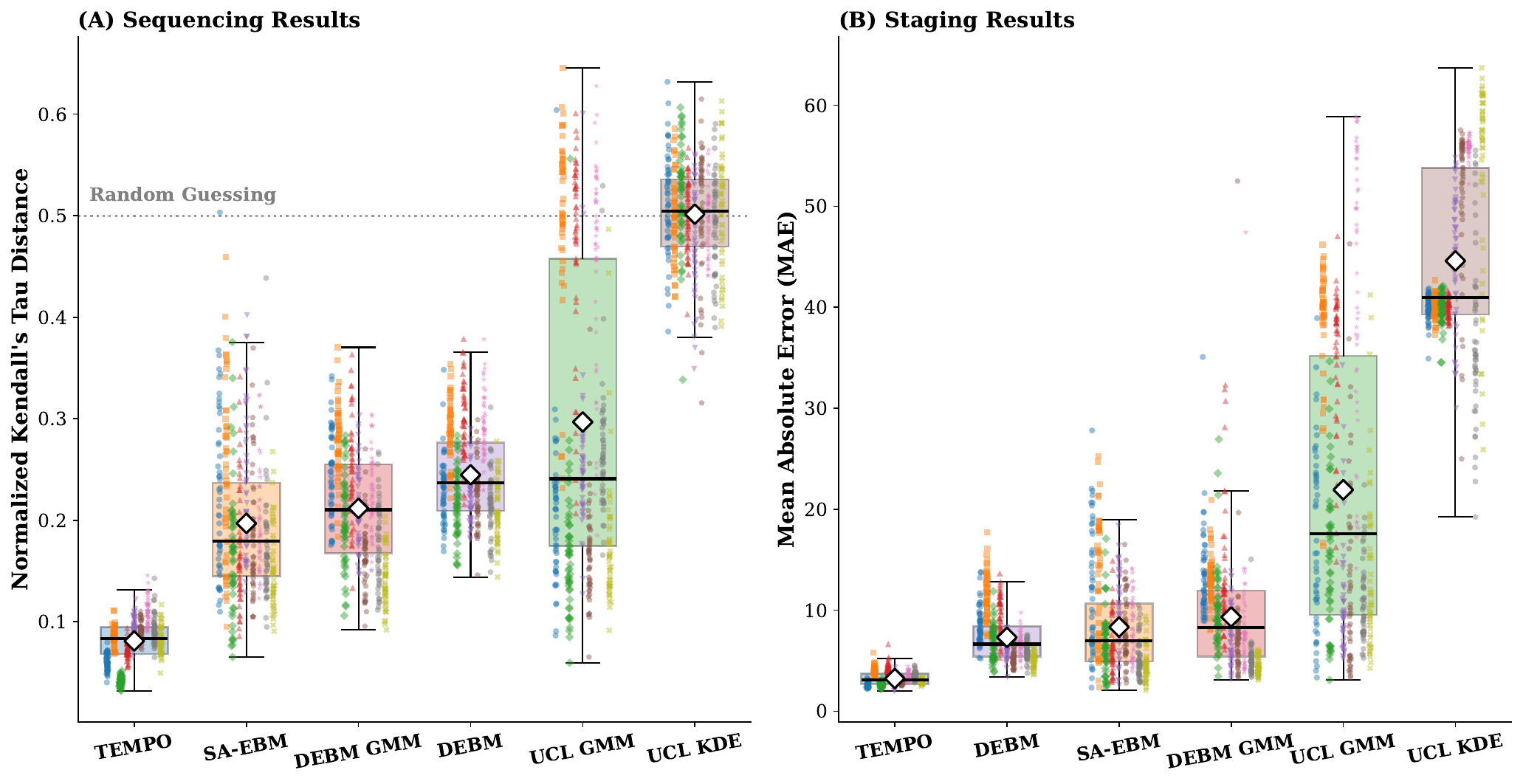} 
    \caption{\textbf{High-Dimensional Performance.} Comparison of \textsc{Tempo} and SA-EBM across (A) sequencing accuracy (Tau Distance) and (B) staging precision (MAE) for $B=100$ biomarkers. Individual results ($N=450$) are displayed as ordered strip-plots (Exp 1 $\to$ 9). \textsc{Tempo} maintains superior performance in the high-dimensional setting.}
    \label{fig:highdim_results}
\end{figure*}

\begin{figure*}[htbp]
    \centering
    \includegraphics[width=1\textwidth]{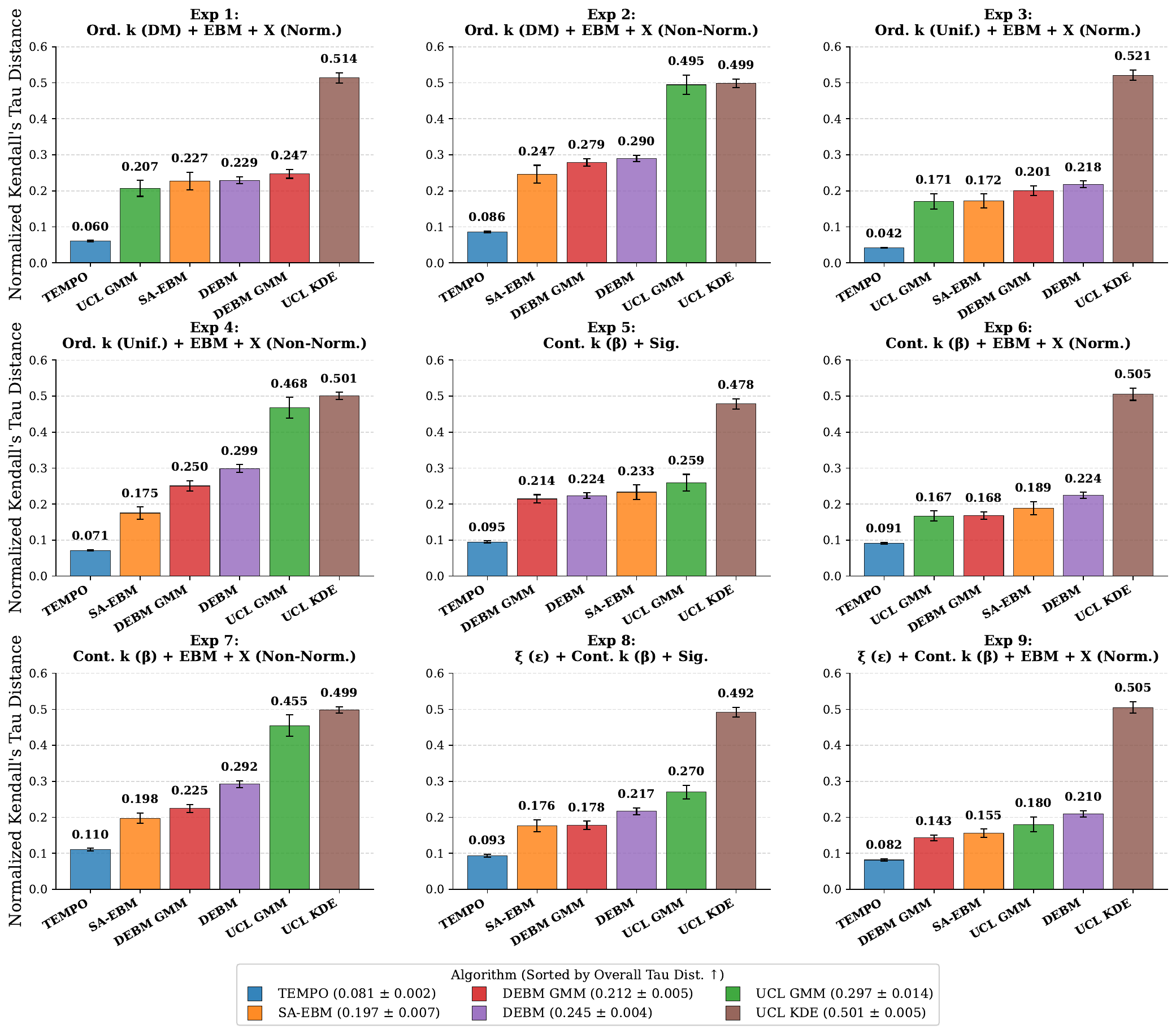} 
    \caption{\textbf{High-Dimensional Event Sequencing Performance ($B=100$).} Comparison of normalized Tau distance between \textsc{Tempo} and SA-EBM across nine generative frameworks. In this high-dimensional setting, \textsc{Tempo} achieves a 58.88\% mean reduction in Tau distance compared to SA-EBM. The results demonstrate that \textsc{Tempo}'s self-attention mechanism on biomarkers effectively captures pathological dependencies even as the feature space scales ten-fold from the low-dimensional study. Error bars represent 95\% confidence intervals across the 50 test datasets in each experiment.}
    \label{fig:highdim_tau_all}
\end{figure*}

\begin{figure*}[htbp]
    \centering
    \includegraphics[width=1\textwidth]{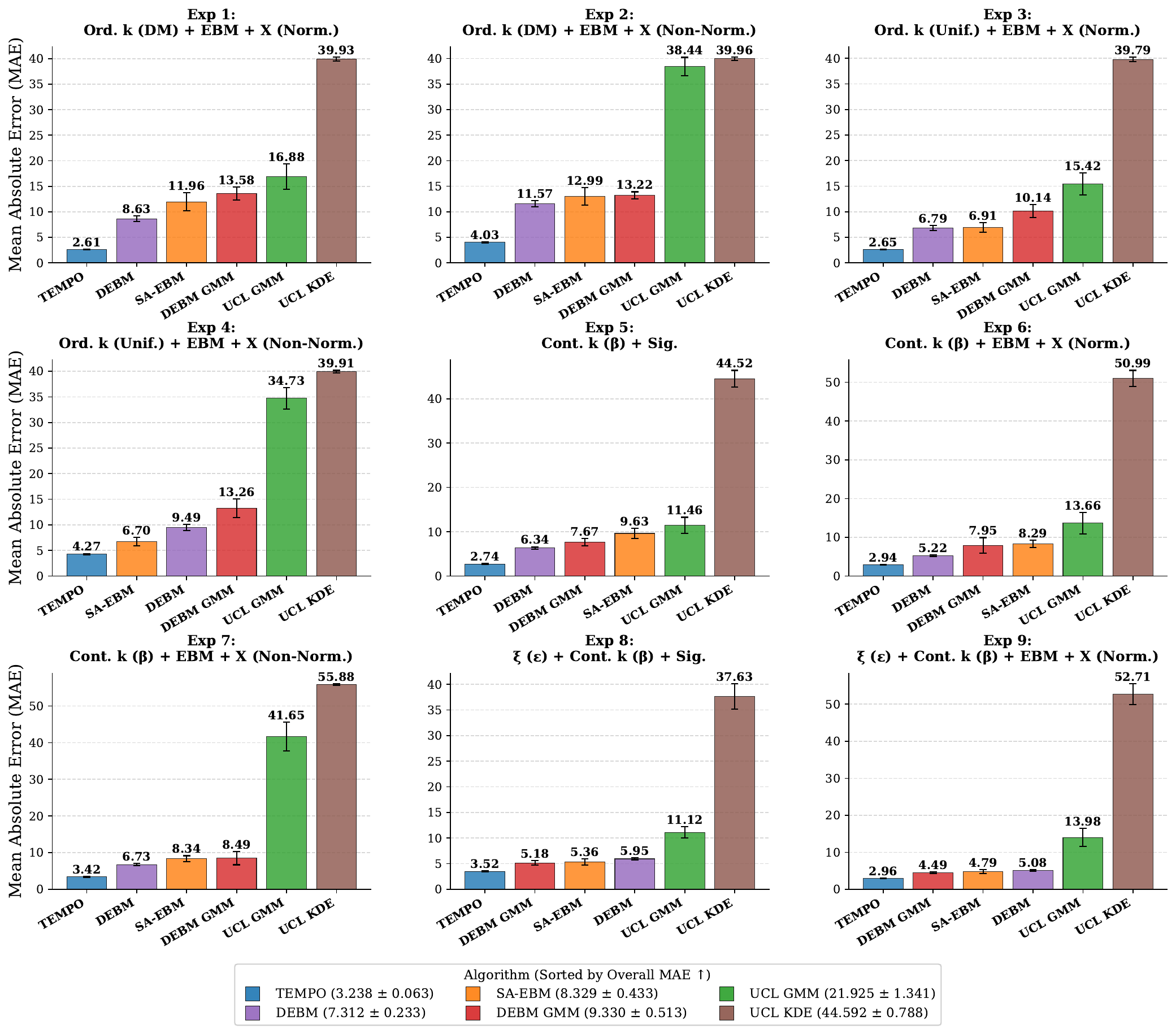} 
    \caption{\textbf{High-Dimensional Patient Staging Performance ($B=100$).} Comparison of staging MAE across nine experimental conditions using the UnifiedTransformer architecture. \textsc{Tempo} achieves a mean MAE reduction of 61.10\% relative to SA-EBM. Notably, when normalized by the number of biomarkers ($MAE/B$), staging error decreases from 5.7\% (low-dimensional, $B=12$) to 3.2\% ($B=100$), confirming that \textsc{Tempo}’s patient-level attention benefits from higher dimensionality. Error bars represent 95\% confidence intervals across the 50 test datasets in each experiment.}
    \label{fig:highdim_mae_all}
\end{figure*}

\begin{table*}[htbp]
\centering
\caption{\textbf{Per-Experiment Performance (High Dimensional, $B=100$).} Mean normalized Tau distance and staging MAE with 95\% confidence intervals. Lower values indicate better performance. \textsc{Tempo} achieves the lowest Tau distance and staging MAE across all nine experiments.}
\label{tab:highdim_per_exp}
\small
\begin{tabular}{@{}lcccc@{}}
\toprule
& \multicolumn{2}{c}{\textbf{Tau Distance} $\downarrow$} & \multicolumn{2}{c}{\textbf{Staging MAE} $\downarrow$} \\
\cmidrule(lr){2-3} \cmidrule(lr){4-5}
\textbf{Exp} & \textsc{Tempo} & SA-EBM & \textsc{Tempo} & SA-EBM \\
\midrule
1 & $\mathbf{0.060 \pm 0.002}$ & $0.227 \pm 0.024$ & $\mathbf{2.61 \pm 0.07}$ & $11.96 \pm 1.76$ \\
2 & $\mathbf{0.086 \pm 0.003}$ & $0.247 \pm 0.024$ & $\mathbf{4.03 \pm 0.11}$ & $12.99 \pm 1.73$ \\
3 & $\mathbf{0.042 \pm 0.001}$ & $0.172 \pm 0.020$ & $\mathbf{2.65 \pm 0.06}$ & $6.91 \pm 0.93$ \\
4 & $\mathbf{0.071 \pm 0.002}$ & $0.175 \pm 0.017$ & $\mathbf{4.27 \pm 0.14}$ & $6.70 \pm 0.82$ \\
5 & $\mathbf{0.095 \pm 0.003}$ & $0.233 \pm 0.020$ & $\mathbf{2.74 \pm 0.13}$ & $9.63 \pm 1.11$ \\
6 & $\mathbf{0.091 \pm 0.002}$ & $0.189 \pm 0.018$ & $\mathbf{2.94 \pm 0.07}$ & $8.29 \pm 0.95$ \\
7 & $\mathbf{0.110 \pm 0.004}$ & $0.198 \pm 0.014$ & $\mathbf{3.42 \pm 0.11}$ & $8.34 \pm 0.83$ \\
8 & $\mathbf{0.093 \pm 0.004}$ & $0.176 \pm 0.017$ & $\mathbf{3.52 \pm 0.10}$ & $5.36 \pm 0.58$ \\
9 & $\mathbf{0.082 \pm 0.004}$ & $0.155 \pm 0.012$ & $\mathbf{2.96 \pm 0.06}$ & $4.79 \pm 0.49$ \\
\bottomrule
\end{tabular}
\end{table*}

\begin{table*}[t]
\centering
\caption{\textbf{Cross-Condition Sequence MAE Generalization. (High Dimensional)} Mean Absolute Error (MAE) between predicted and ground-truth event times across training and testing environments. Diagonal entries (bold) represent in-domain performance, while off-diagonal entries indicate generalization capability across disparate pathological models.}
\label{tab:sequence_mae_highdim}
\resizebox{\textwidth}{!}{
\begin{tabular}{lcccccccccc}
\toprule
\textbf{Trained \textbackslash Tested} & \textbf{Exp 1} & \textbf{Exp 2} & \textbf{Exp 3} & \textbf{Exp 4} & \textbf{Exp 5} & \textbf{Exp 6} & \textbf{Exp 7} & \textbf{Exp 8} & \textbf{Exp 9} & \textbf{Row Mean} \\
\midrule
\textbf{Exp 1} & \textbf{5.335} & 16.987 & 4.213 & 16.876 & 20.990 & 13.199 & 19.341 & 20.305 & 11.830 & 14.342 \\
\textbf{Exp 2} & 8.399 & \textbf{7.758} & 6.153 & 6.496 & 28.252 & 14.242 & 13.608 & 26.303 & 12.429 & 13.738 \\
\textbf{Exp 3} & 6.981 & 17.243 & \textbf{3.321} & 17.263 & 18.485 & 13.246 & 18.592 & 18.044 & 11.974 & 13.905 \\
\textbf{Exp 4} & 8.721 & 7.825 & 6.876 & \textbf{5.396} & 26.587 & 12.402 & 13.369 & 25.214 & 10.283 & 12.964 \\
\textbf{Exp 5} & 12.047 & 17.510 & 10.154 & 17.867 & \textbf{8.687} & 11.819 & 17.741 & 8.653 & 11.516 & 12.888 \\
\textbf{Exp 6} & 10.445 & 16.742 & 10.063 & 17.760 & 18.815 & \textbf{9.570} & 17.470 & 17.712 & 10.466 & 14.338 \\
\textbf{Exp 7} & 10.034 & 9.700 & 10.290 & 9.354 & 25.651 & 10.250 & \textbf{10.424} & 24.301 & 10.177 & 13.353 \\
\textbf{Exp 8} & 12.027 & 16.592 & 10.890 & 16.817 & 9.329 & 10.498 & 16.954 & \textbf{9.298} & 9.965 & 12.485 \\
\textbf{Exp 9} & 10.782 & 17.226 & 13.159 & 18.769 & 25.144 & 8.640 & 17.856 & 23.254 & \textbf{7.819} & 15.850 \\
\midrule
\textbf{Col Mean} & 9.419 & 14.176 & 8.347 & 14.066 & 20.216 & 11.541 & 16.151 & 19.231 & 10.718 & \textbf{13.763} \\
\bottomrule
\end{tabular}
}
\end{table*}

\begin{table*}[t]
\centering
\caption{\textbf{Cross-Condition Tau Distance Generalization (High Dimensional).} Average normalized Tau distance across training and testing environments for $B=100$ biomarkers. Diagonal entries (bold) represent in-domain performance. This matrix illustrates the scalability of TEMPO's sequencing logic in high-dimensional feature spaces.}
\label{tab:highdim_cross_tau}
\resizebox{\textwidth}{!}{
\begin{tabular}{lcccccccccc}
\toprule
\textbf{Trained \textbackslash Tested} & \textbf{Exp 1} & \textbf{Exp 2} & \textbf{Exp 3} & \textbf{Exp 4} & \textbf{Exp 5} & \textbf{Exp 6} & \textbf{Exp 7} & \textbf{Exp 8} & \textbf{Exp 9} & \textbf{Row Mean} \\
\midrule
\textbf{Exp 1} & \textbf{0.060} & 0.250 & 0.054 & 0.259 & 0.278 & 0.112 & 0.245 & 0.313 & 0.166 & 0.193 \\
\textbf{Exp 2} & 0.077 & \textbf{0.086} & 0.079 & 0.086 & 0.385 & 0.125 & 0.129 & 0.396 & 0.167 & 0.170 \\
\textbf{Exp 3} & 0.060 & 0.253 & \textbf{0.042} & 0.264 & 0.263 & 0.108 & 0.258 & 0.304 & 0.131 & 0.187 \\
\textbf{Exp 4} & 0.101 & 0.092 & 0.095 & \textbf{0.071} & 0.353 & 0.134 & 0.128 & 0.377 & 0.159 & 0.168 \\
\textbf{Exp 5} & 0.119 & 0.240 & 0.115 & 0.259 & \textbf{0.095} & 0.152 & 0.251 & 0.099 & 0.172 & 0.167 \\
\textbf{Exp 6} & 0.078 & 0.228 & 0.088 & 0.254 & 0.266 & \textbf{0.091} & 0.243 & 0.310 & 0.097 & 0.184 \\
\textbf{Exp 7} & 0.123 & 0.111 & 0.139 & 0.123 & 0.349 & 0.107 & \textbf{0.110} & 0.374 & 0.109 & 0.172 \\
\textbf{Exp 8} & 0.100 & 0.227 & 0.077 & 0.239 & 0.094 & 0.141 & 0.248 & \textbf{0.093} & 0.153 & 0.152 \\
\textbf{Exp 9} & 0.064 & 0.225 & 0.048 & 0.250 & 0.381 & 0.085 & 0.259 & 0.396 & \textbf{0.082} & 0.199 \\
\midrule
\textbf{Col Mean} & 0.087 & 0.190 & 0.082 & 0.201 & 0.274 & 0.117 & 0.208 & 0.296 & 0.137 & \textbf{0.177} \\
\bottomrule
\end{tabular}
}
\end{table*}

\begin{table*}[t]
\centering
\caption{\textbf{Cross-Condition Staging MAE Generalization (High Dimensional).} Mean Absolute Error (MAE) between predicted and ground-truth patient stages for $B=100$. Diagonal entries (bold) represent in-domain performance. Values show how TEMPO’s staging accuracy scales favorably with dimensionality when compared to the low-dimensional results.}
\label{tab:highdim_cross_mae}
\resizebox{\textwidth}{!}{
\begin{tabular}{lcccccccccc}
\toprule
\textbf{Trained \textbackslash Tested} & \textbf{Exp 1} & \textbf{Exp 2} & \textbf{Exp 3} & \textbf{Exp 4} & \textbf{Exp 5} & \textbf{Exp 6} & \textbf{Exp 7} & \textbf{Exp 8} & \textbf{Exp 9} & \textbf{Row Mean} \\
\midrule
\textbf{Exp 1} & \textbf{2.608} & 4.258 & 3.148 & 6.156 & 20.647 & 4.671 & 8.421 & 27.058 & 6.429 & 9.266 \\
\textbf{Exp 2} & 3.352 & \textbf{4.030} & 3.591 & 4.913 & 28.295 & 4.575 & 7.418 & 32.497 & 5.764 & 10.493 \\
\textbf{Exp 3} & 3.856 & 5.206 & \textbf{2.649} & 4.882 & 32.982 & 3.470 & 7.414 & 25.192 & 4.400 & 10.006 \\
\textbf{Exp 4} & 4.897 & 5.375 & 3.738 & \textbf{4.270} & 31.514 & 4.465 & 5.922 & 34.920 & 4.109 & 11.023 \\
\textbf{Exp 5} & 10.522 & 18.946 & 9.770 & 19.251 & \textbf{2.741} & 3.536 & 7.212 & 5.722 & 5.664 & 9.263 \\
\textbf{Exp 6} & 11.140 & 12.187 & 10.951 & 13.377 & 13.376 & \textbf{2.942} & 4.948 & 17.823 & 5.972 & 10.302 \\
\textbf{Exp 7} & 9.180 & 9.963 & 9.199 & 10.053 & 22.765 & 2.709 & \textbf{3.416} & 24.698 & 4.433 & 10.713 \\
\textbf{Exp 8} & 8.521 & 21.790 & 7.700 & 21.364 & 3.948 & 4.798 & 10.612 & \textbf{3.522} & 4.289 & 9.616 \\
\textbf{Exp 9} & 5.078 & 8.025 & 4.213 & 7.049 & 47.156 & 2.824 & 4.381 & 47.649 & \textbf{2.959} & 14.370 \\
\midrule
\textbf{Col Mean} & 6.573 & 9.975 & 6.107 & 10.146 & 22.603 & 3.777 & 6.638 & 24.342 & 4.891 & \textbf{10.561} \\
\bottomrule
\end{tabular}
}
\end{table*}

\begin{figure*}[htbp]
    \centering
    \includegraphics[width=0.85\textwidth]{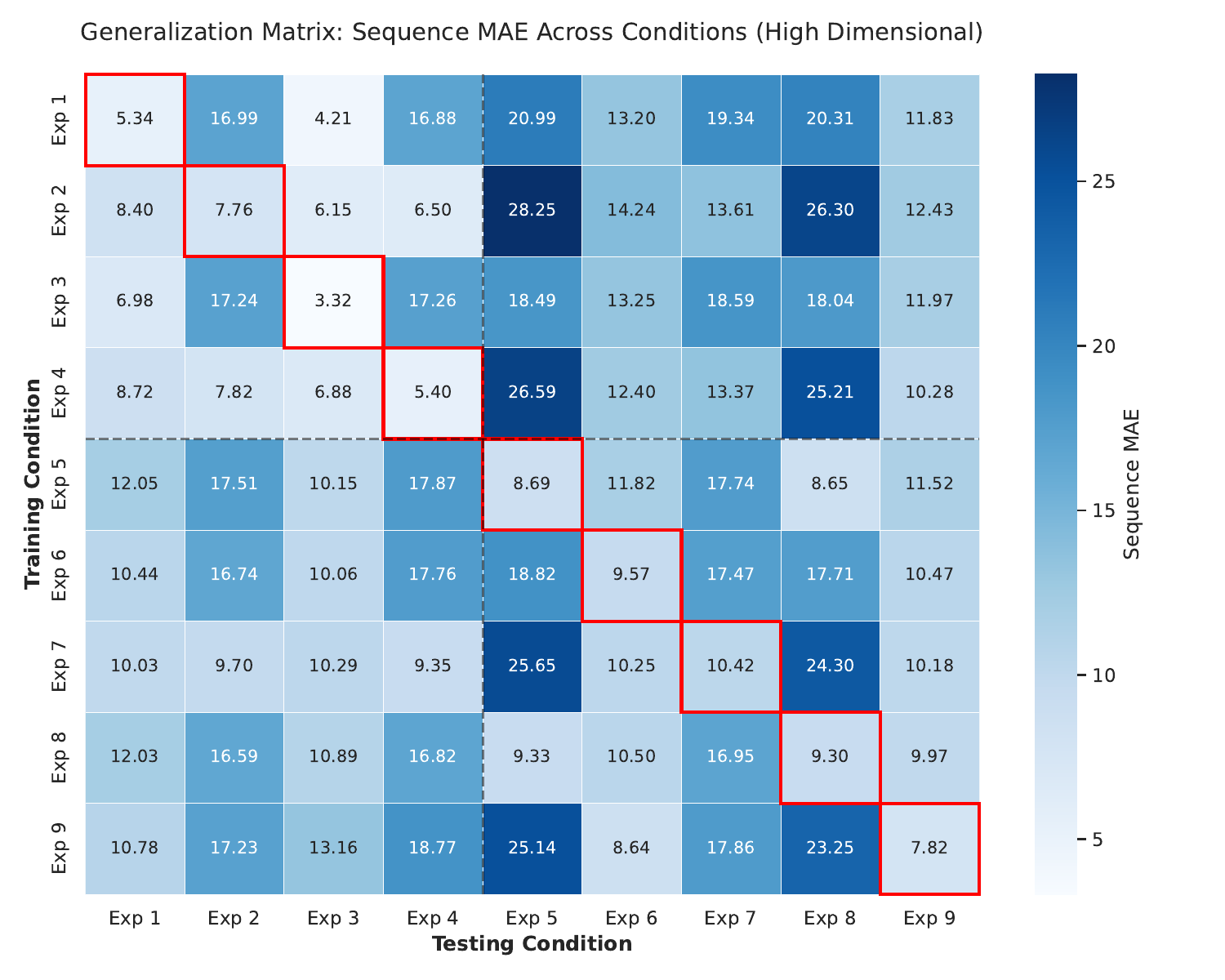} 
    \caption{\textbf{Cross-Condition Sequence Generalization (High Dimensional).} Heatmap of Sequence MAE across 81 training and testing permutations ($B=100$). Red squares along the diagonal denote in-domain performance. Columns 5 and 8 (Sigmoid-generated test data with continuous event times) exhibit higher col-mean errors due to scale mismatch with discrete-event-time experiments. Models trained on Exp~8 (Sigmoid with continuous event times) achieve the lowest row-mean error ($12.485$), demonstrating that Sigmoid training with continuous event times improves cross-experiment generalization.}
    \label{fig:sequence_mae_highdim}
\end{figure*}

\clearpage
\section{ADNI}
\label{apd:adni}

We included only baseline visits (VISCODE = bl) from participants diagnosed as cognitively normal (CN), early or late mild cognitive impairment (EMCI/LMCI), or Alzheimer's disease (AD). We selected twelve biomarkers commonly reported in prior EBM studies: cognitive assessments (MMSE, ADAS13, RAVLT\_immediate), cerebrospinal fluid markers (PTAU, TAU, ABETA), and MRI-derived volumetric measures (Ventricles, WholeBrain, MidTemp, Fusiform, Entorhinal, Hippocampus). Participants with missing values on any biomarker were excluded, yielding a final cohort of $N = 726$ participants with 152 controls (21\%).

\begin{figure}[htbp]
    \centering
    \includegraphics[width=0.5\columnwidth]{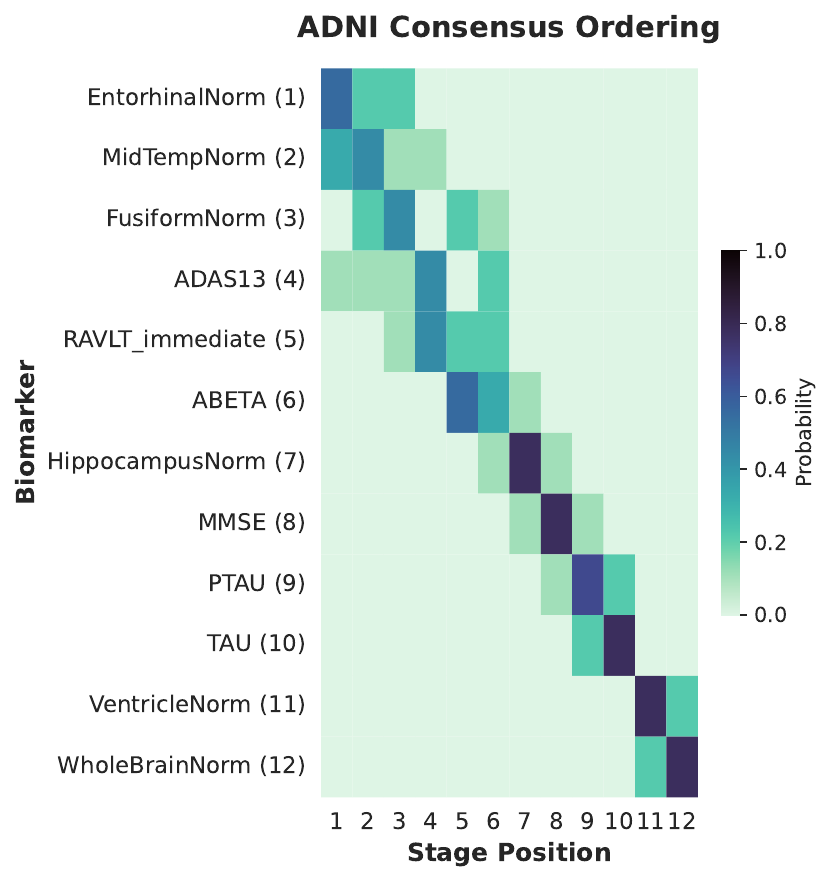}
    \caption{\textbf{ADNI Biomarker Progression.} Positional variance diagram showing the temporal ordering of 12 biomarkers. Cell probabilities are calculated based on the frequency of each biomarker appearing at each stage position across the nine trained models.}
    \label{fig:adni_heatmap}
\end{figure}

\begin{table*}[htbp]
\centering
\caption{Biomarker Event Ordering across Nine Experimental Iterations. Ranks represent the sequence of abnormality (1 = earliest). Mean, Standard Deviation (Std), and 95\% Confidence Intervals (CI) are calculated across all experiments.}
\label{tab:biomarker_position_matrix}
\small 
\begin{tabular}{@{}lccccccccc rrc@{}}
\toprule
\textbf{Biomarker} & \textbf{e1} & \textbf{e2} & \textbf{e3} & \textbf{e4} & \textbf{e5} & \textbf{e6} & \textbf{e7} & \textbf{e8} & \textbf{e9} & \textbf{Mean} & \textbf{Std} & \textbf{95\% CI} \\ 
\midrule
EntorhinalNorm & 1 & 3 & 1 & 1 & 2 & 3 & 2 & 1 & 1 & 1.7 & 0.9 & [1.0, 2.3] \\
MidTempNorm & 2 & 2 & 2 & 4 & 1 & 1 & 1 & 3 & 2 & 2.0 & 1.0 & [1.2, 2.8] \\
FusiformNorm & 3 & 5 & 6 & 5 & 3 & 2 & 3 & 2 & 3 & 3.6 & 1.4 & [2.5, 4.7] \\
ADAS13 & 4 & 1 & 3 & 2 & 6 & 4 & 4 & 6 & 4 & 3.8 & 1.6 & [2.5, 5.0] \\
RAVLT\_immediate & 5 & 4 & 4 & 3 & 4 & 6 & 5 & 4 & 6 & 4.6 & 1.0 & [3.8, 5.3] \\
ABETA & 6 & 6 & 5 & 6 & 5 & 5 & 7 & 5 & 5 & 5.6 & 0.7 & [5.0, 6.1] \\
HippocampusNorm & 7 & 8 & 7 & 7 & 7 & 7 & 6 & 7 & 7 & 7.0 & 0.5 & [6.6, 7.4] \\
MMSE & 8 & 7 & 8 & 8 & 8 & 8 & 8 & 8 & 9 & 8.0 & 0.5 & [7.6, 8.4] \\
PTAU & 9 & 10 & 9 & 9 & 9 & 9 & 10 & 9 & 8 & 9.1 & 0.6 & [8.6, 9.6] \\
TAU & 10 & 9 & 10 & 10 & 10 & 10 & 9 & 10 & 10 & 9.8 & 0.4 & [9.4, 10.1] \\
VentricleNorm & 11 & 11 & 11 & 11 & 12 & 11 & 11 & 12 & 11 & 11.2 & 0.4 & [10.9, 11.6] \\
WholeBrainNorm & 12 & 12 & 12 & 12 & 11 & 12 & 12 & 11 & 12 & 11.8 & 0.4 & [11.4, 12.1] \\
\bottomrule
\end{tabular}
\end{table*}

\begin{figure*}[t]
    \centering
    \includegraphics[width=0.95\textwidth]{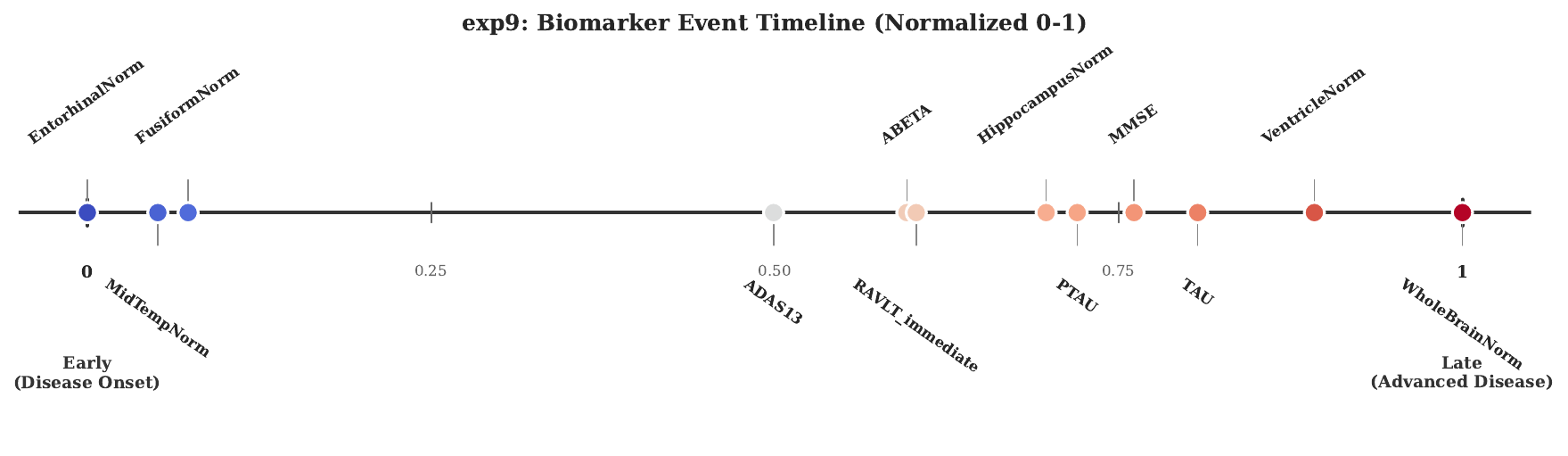}
    \caption{\textbf{Continuous Temporal Spacing of Biomarker Abnormality (Exp 9).} The timeline is generated by the Exp 9 model (EBM measurement model with continuous event times). The horizontal axis is min-max normalized to $[0,1]$. The overall pattern is similar to Exp 8 (Fig.~\ref{fig:continuous_timeline}).}
    \label{fig:continuous_timeline_exp9}
\end{figure*}

\begin{table}[htbp]
\centering
\caption{\textbf{Average Predicted Stage by Diagnosis Group.} Mean ordinal stage (out of 12) predicted by \textsc{Tempo} for each ADNI diagnosis group, averaged across all nine experimental models.}
\label{tab:staging_by_dx}
\small
\begin{tabular}{lrrrr}
\toprule
Model & CN & EMCI & LMCI & AD \\
\midrule
Exp 1 & 0.01 & 5.28 & 9.35 & 11.56 \\
Exp 2 & 0.01 & 4.88 & 7.95 & 10.14 \\
Exp 3 & 0.00 & 5.12 & 8.79 & 11.06 \\
Exp 4 & 0.01 & 4.38 & 7.74 & 10.53 \\
Exp 5 & 0.00 & 6.11 & 8.06 & 10.14 \\
Exp 6 & 0.49 & 6.42 & 8.65 & 10.01 \\
Exp 7 & $-$0.01 & 6.99 & 8.36 & 9.55 \\
Exp 8 & 0.00 & 5.56 & 8.64 & 11.50 \\
Exp 9 & 0.00 & 4.91 & 9.12 & 11.58 \\
\midrule
\textbf{Mean} & \textbf{0.06} & \textbf{5.52} & \textbf{8.52} & \textbf{10.68} \\
\bottomrule
\end{tabular}
\end{table}

\clearpage
\section{ADNI Information}
\label{apd:adniinfo}


A complete listing of ADNI
investigators can be found at:
\url{http://adni.loni.usc.edu/wp-content/uploads/how_to_apply/ADNI_Acknowledgement_List.pdf}

\end{document}